\documentclass[runningheads]{llncs}
\usepackage{graphicx}

\usepackage{booktabs} 
\usepackage{amssymb}
\usepackage{amsmath}

\usepackage[table]{xcolor}
\usepackage{multirow}
\usepackage{algorithm}
\usepackage{algorithmic}

\newtheorem{rem}{Remark}

\usepackage{subfig}
\usepackage{adjustbox}

\usepackage{hyperref}
\begin{document}
\title{Data Augmentation with Variational Autoencoders and Manifold Sampling}
%
%
\author{Clément Chadebec and Stéphanie Allassonnière}
\authorrunning{Chadebec and Allassonnière}
%
\institute{Université de Paris, INRIA, Centre de recherche des Cordeliers, INSERM, Sorbonne Université, Paris, France\\\email{\{clement.chadebec, stephanie.allassonniere\}@inria.fr}}

\maketitle              
\begin{abstract}
    We propose a new efficient way to sample from a Variational Autoencoder in the challenging low sample size setting\footnote{A code is available at \url{https://github.com/clementchadebec/Data_Augmentation_with_VAE-DALI}}. This method reveals particularly well suited to perform data augmentation in such a low data regime and is validated across various standard and \emph{real-life} data sets. In particular, this scheme allows to greatly improve classification results on the OASIS database where balanced accuracy jumps from 80.7\% for a classifier trained with the raw data to 88.6\% when trained only with the synthetic data generated by our method. Such results were also observed on 3 standard data sets and with other classifiers.

\end{abstract}
\section{Introduction}
\label{Sec: Introduction}

Despite the apparent availability of always bigger data sets, the lack of data remains a key issue for many fields of application. One of them is medicine where practitioners have to deal with potentially very high dimensional data (\emph{e.g.} functional Magnetic Resonance Imaging for neuroimaging) along with very low sample sizes (\emph{e.g.} rare diseases or heterogeneous cancers) which make statistical analysis challenging and unreliable. In addition, the wide use of algorithms heavily relying on the deep learning framework \cite{goodfellow_deep_2016} and requiring a large amount of data has made the need for data augmentation (DA) crucial to avoid poor performance or over-fitting \cite{shorten_survey_2019}. As an example, a classic way to perform DA on images consists in applying simple transformations such as adding random noise, rotations etc. However, it may be easily understood that such augmentation techniques are strongly data dependent\footnote{Think of digits where rotating a \emph{6} gives a \emph{9} for example.} and may still require the intervention of an expert assessing the relevance of the augmented samples. The recent development of generative models such as Generative Adversarial Networks (GAN) \cite{goodfellow_generative_2014} or Variational AutoEncoders (VAE) \cite{kingma_auto-encoding_2014,rezende_stochastic_2014} paves the way for consideration of another way to augment the training data. While GANs have already seen some success \cite{lintas_biomedical_2017,zhu_data_2017,antoniou_data_2018} and even for medical data \cite{liu_wasserstein_2019,sandfort_data_2019} VAEs have been of least interest. One limitation of the use of both generative models relies in their need of a large amount of data to be able to generate faithfully. 
In this paper, we argue that VAEs can actually be used to perform DA in challenging contexts provided that we amend the way we generate the data. Hence, we propose:
    \begin{itemize}
        \item A new non \emph{prior-dependent} generation method using the learned geometry of the latent space and consisting in exploring it by sampling along geodesics.
        \item To use this method to perform DA in the small sample size setting on  standard data sets and real data from OASIS database \cite{marcus_open_2007} where it allows to remarkably improve classification results.
    \end{itemize}

\section{Variational Autoencoder}

Given a set of data $x \in \mathcal{X}$, a VAE aims at maximizing the likelihood of the associated parametric model $\{\mathbb{P}_{\theta}, \theta \in \Theta\}$. Assuming that there exist latent variables $z \in \mathcal{Z}$ living in a lower dimensional space $\mathcal{Z}$, the marginal distribution writes
\begin{equation}\label{Eq:Objective}
    p_{\theta}(x) = \int \limits _{\mathcal{Z}} p_{\theta}(x|z)q(z) dz \,,
\end{equation}
where $q$ is a prior distribution over the latent variables and $p_{\theta}(x|z)$ is most of the time a simple distribution and is referred to as the \emph{decoder}. A variational distribution $q_{\varphi}$ (often taken as Gaussian) aiming at approximating the true posterior distribution and referred to as the \emph{encoder}  is then introduced. Using Importance Sampling allows to derive an unbiased estimate of $p_{\theta}(x)$ such that $ \mathbb{E}_{z \sim q_{\varphi}}\big[\hat{p}_{\theta}\big] = p_{\theta}(x)$. Therefore, a lower bound on the logarithm of the objective function of Eq.~\eqref{Eq:Objective} can be derived using Jensen's inequality:
 \begin{equation}
     \begin{aligned}
      \log p_{\theta}(x) 
                        \geq \mathbb{E}_{z \sim q_{\varphi}}\big[ \log p_{\theta}(x, z) - \log q_{\varphi}(z|x) \big]=ELBO \,.
     \end{aligned}
 \end{equation}
Using the reparametrization trick makes the ELBO tractable and so can be optimised with respect to both $\theta$ and $\varphi$, the \emph{encoder} and \emph{decoder} parameters. Once the model is trained, the decoder acts as a generative model and new data can be generated by simply drawing a sample using the prior $q$ and feeding it to the decoder. Several axes of improvement of this model were recently explored. One of them consists in trying to bring geometry into the model by learning the latent structure of the data seen as a Riemannian manifold \cite{arvanitidis_latent_2018,chadebec_geometry-aware_2020}.

\section{Some Elements on Riemannian Geometry}

In the framework of differential geometry, one may define a Riemannian manifold $\mathcal{M}$ as a smooth manifold endowed with a Riemannian metric $\mathbf{G}$ which is a smooth inner product $\mathbf{G}: p \to \langle \cdot | \cdot \rangle_p$ on the tangent space $T_p\mathcal{M}$ defined at each point $p$ of the manifold.  The length of a curve $\gamma$ between two points of the manifold $z_1, z_2 \in \mathcal{M}$ and parametrized by $ t \in [0, 1]$ such that $\gamma(0) = z_1$ and $\gamma(1) = z_2$ is given by $
    \mathcal{L}(\gamma) = \int \limits _0 ^1 \lVert \dot{\gamma}(t) \rVert_{\gamma(t)} dt = \int \limits _0 ^1 \sqrt{\langle \dot{\gamma}(t) | \dot{\gamma}(t) \rangle_{\gamma(t)}} dt\,.$ Curves minimizing such a length are called geodesics.
For any $p \in \mathcal{M}$,  the exponential map at $p$, $
    \mathrm{Exp}_{p}$, maps a  vector $v$ of the tangent space $T_p\mathcal{M}$ to a point of the manifold $\widetilde{p} \in \mathcal{M}$ such that the geodesic starting at $p$ with initial velocity $v$ reaches $\widetilde{p}$ at time 1. In particular, if the manifold is \emph{geodesically complete}, then $\mathrm{Exp}_p$ is defined on the entire tangent space $T_p\mathcal{M}$.


\section{The Proposed Method}
    
    We propose a new sampling method exploiting the structure of the latent space seen as a Riemannian manifold and independent from the choice of the prior distribution. The view we adopt is to consider the VAE as a tool to perform dimensionality reduction by extracting the latent structure of the data within a lower dimensional space. Having learned such a structure, we propose to exploit it to enhance the data generation process. This differs from the fully probabilistic view which uses the prior to generate. We believe that this is far from being optimal since the prior appears quite strongly data dependent. We will adopt the same setting as \cite{chadebec_geometry-aware_2020} and so use a RHVAE since the metric used by the authors is easily computable, constraints geodesic path to travel through most populated areas of the latent space and the learned Riemannian manifold is geodesically complete. Nonetheless, the proposed method can be used with different metrics as well as long as the exponential map remains computable. We now assume that we are given a latent space with a Riemannian structure where the metric has been estimated from the input data.

\subsection{The Wrapped Normal Distribution}\label{Sec: Wrapped Normal}

\begin{figure}[!t]
\centering
\subfloat{\includegraphics[width=1.5in]{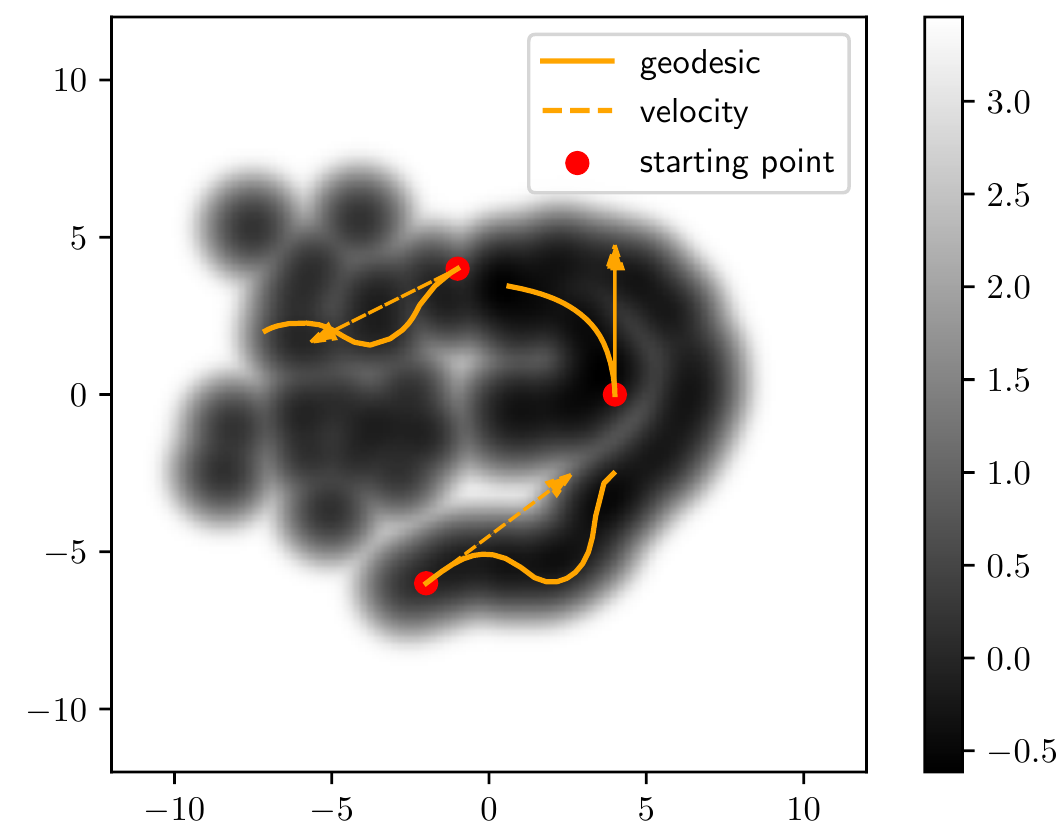}
\label{Geodesic shapes}}
\hfil
\subfloat{\includegraphics[width=1.5in]{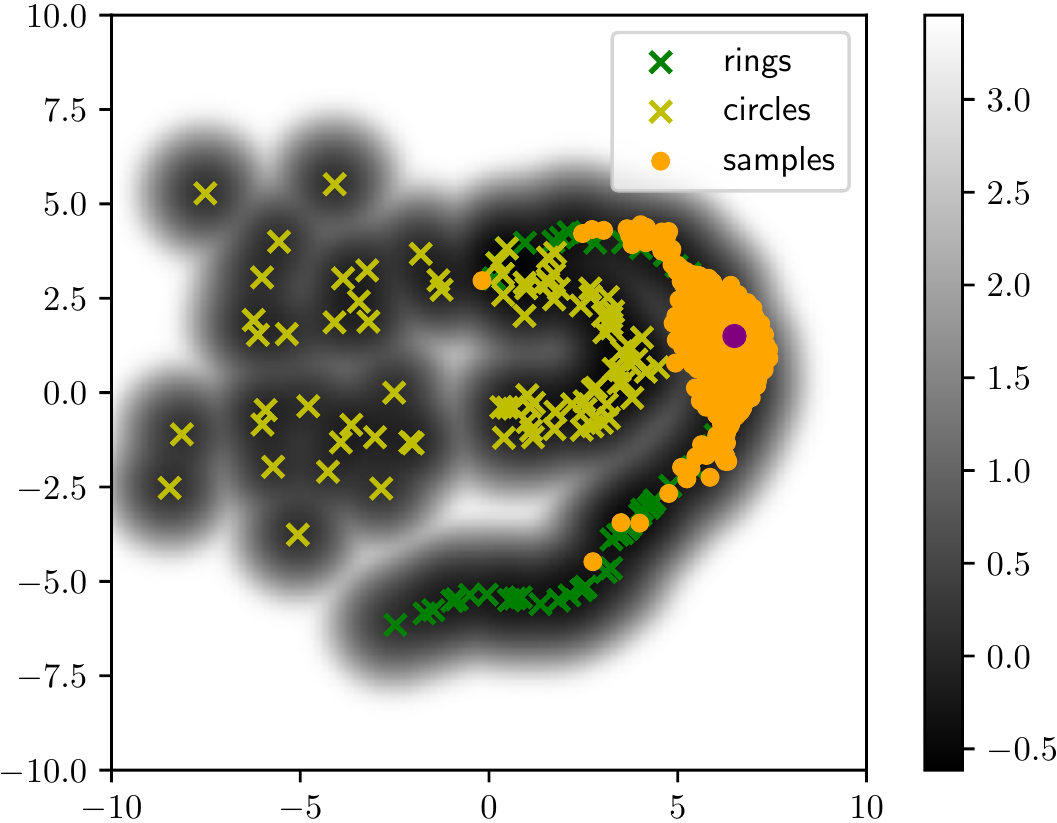}}
\hfil
\subfloat{\includegraphics[width=1.5in]{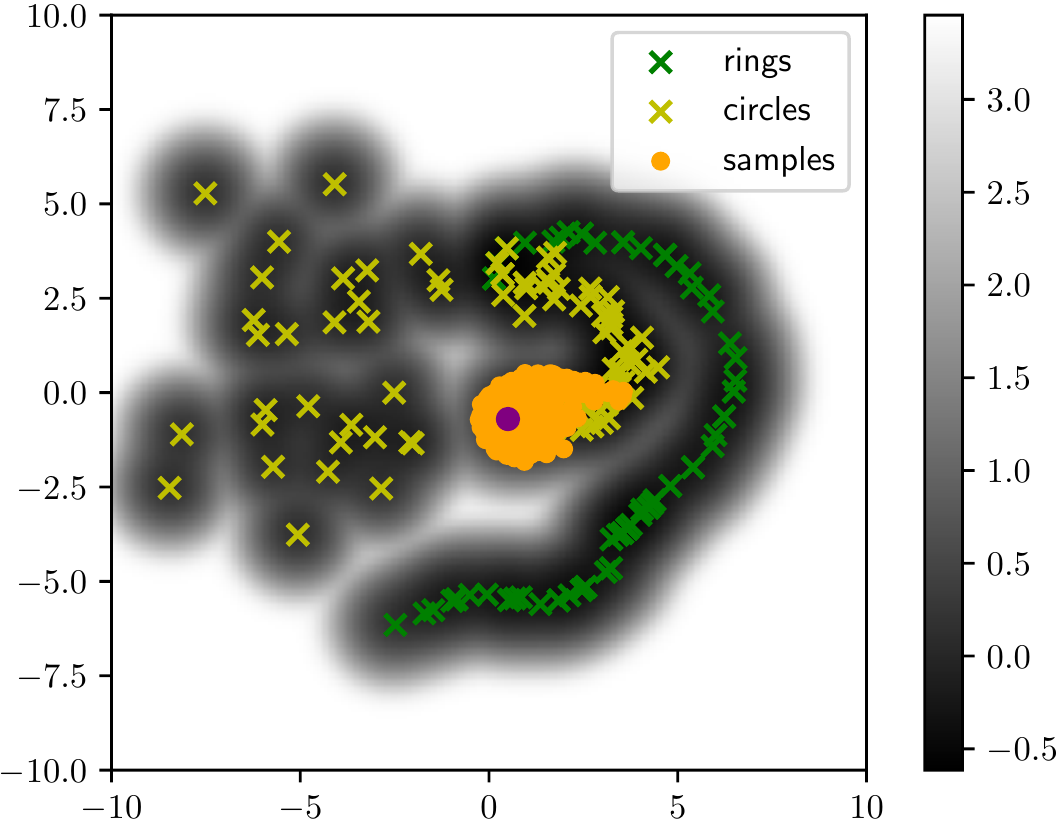}
\label{Geodesic Fashion}}
\caption{\emph{Left}: Geodesic \emph{shooting} in a latent space learned by a RHVAE with different starting points (red dots) and initial velocities (orange arrows). \emph{Middle and right}: Samples from the wrapped normal $\mathcal{N}^{W}(p, I_d)$. The log metric volume element $\log \sqrt{\det \mathbf{G}(z)}$ is presented in gray scale. }
\label{Fig: Geodesic shooting}
\end{figure}

The notion of normal distribution may be extended to Riemannian manifolds in several ways. One of them is the \emph{wrapped} normal distribution. The main idea is to define a classic normal distribution $\mathcal{N}(0, \Sigma)$ on the tangent space $T_p\mathcal{M}$ for any $p \in \mathcal{M}$ and pushing it forward to the manifold using the exponential map. This defines a probability distribution on the manifold $\mathcal{N}^W(p, \Sigma)$ called the \emph{wrapped} normal distribution. Sampling from this distribution is straight forward and consists in drawing a velocity in the tangent space from  $\mathcal{N}(0, \Sigma)$ and mapping it onto the manifold using the exponential map \cite{mallasto_wrapped_2018}. Hence, the \emph{wrapped} normal allows for a latent space prospecting along geodesic paths. Nonetheless, this requires to compute $\mathrm{Exp}_{p}$ which can be performed with a numerical scheme (see. App.~C). On the left of Fig.~\ref{Fig: Geodesic shooting} are displayed some geodesic paths with respect to the metric and different starting points (red dots) and initial velocities (orange arrows). Samples from $\mathcal{N}^W(p, I_d)$ are also presented in the middle and the right along with the encoded input data. As expected this distribution takes into account the local geometry of the manifold thanks to the geodesic shooting steps. This is a very interesting property since it encourages the samples to remain close to the data as geodesics tend to travel through locations with the lowest volume element $\sqrt{\det \mathbf{G}(z)}$ and so avoid areas with very poor information.

\subsection{Riemannian Random Walk}\label{Sec: Riemannian RW}
A natural way to explore the latent space of a VAE consists in using a random walk like algorithm which moves from one location to another with a certain probability. The idea here is to create a \emph{geometry-aware} Markov Chain ($z^t)_{t\in\mathbb{N}}$ where $z^{t+1}$ is sampled using the \emph{wrapped} normal $z^{t+1} \sim \mathcal{N}^W(z^t, \Sigma)$. However, a drawback of such a method is that every sample of the chain is accepted regardless of its relevance. Nonetheless, by design, the learned metric is such that it has a high volume element far from the data \cite{chadebec_geometry-aware_2020}. This implies that it encodes in a way the amount of information contained at a specific location of the latent space. The higher the volume element, the less information we have. The same idea was used in \cite{lebanon_metric_2006} where the author proposed to see the inverse metric volume element as a maximum likelihood objective to perform metric learning. In our case the likelihood definition writes
 \begin{equation}\label{Eq: Likelihoog metric}
    \mathcal{L}(z) = \frac{\rho_S(z)\sqrt{\det \mathbf{G}^{-1}(z)}}{\int \limits_{\mathbb{R}^d}\rho_S(z)\sqrt{\det \mathbf{G}^{-1}(z)}dz}\,,
\end{equation}
where $\rho_S(z)=1$ if $z \in S$, $0$ otherwise, and $S$ is taken as a compact set so that the integral is well defined. Hence, we propose to use this measure to assess the samples quality as an \emph{acceptance-rejection} rate $\alpha$ in the chain where
$
    \alpha(\widetilde{z}, z) = \min \Bigg(1, \frac{\sqrt{\det \mathbf{G}^{-1}(\widetilde{z})}}{\sqrt{\det \mathbf{G}^{-1}(z)}}\Bigg)\,,
$ $z$ is the current state of the chain and $\widetilde{z}$ is the proposal obtained by sampling from the \emph{wrapped} Gaussian $\mathcal{N}^{W}(z, \Sigma)$. The idea is to compare the relevance of the proposed sample to the current one. The ratio is such that any new sample improving the likelihood metric $\mathcal{L}$ is automatically accepted while a sample degrading the measure is more likely to be rejected in the spirit of Hasting-Metropolis sampler. A pseudo-code is provided in Alg.~\ref{Alg:HM - Riemannian Random Walk}.
\begin{algorithm}[ht]
     \caption{Riemannian random walk}
     \label{Alg:HM - Riemannian Random Walk}
    \begin{algorithmic}
    \STATE{\bfseries Input:} $z_0$, $\Sigma$
     \FOR{$t = 1 \to T$}
     \STATE Draw $v_t \sim \mathcal{N}(0, \Sigma)$\;
     \STATE $\widetilde{z_t} \leftarrow \mathrm{Exp}_{z_{t-1}}(v_t)$ \; 
     \STATE Accept the proposal $\widetilde{z_t}$ with probability $\alpha$\; 
     \ENDFOR
    \end{algorithmic}
\end{algorithm}
\subsection{Discussion}
It may be easily understood that the choice of the covariance matrix $\Sigma$ in Alg.~\ref{Alg:HM - Riemannian Random Walk} has quite an influence on the resulting sampling. On the one hand, a $\Sigma$ with strong eigenvalues will imply drawing velocities of potentially high magnitude allowing for a better prospecting but proposals are more likely to be rejected. On the other hand, small eigenvalues involve a high acceptance rate but it will take longer to prospect the manifold. An adaptive method where $\Sigma$ depends on $\mathbf{G}$ may be considered and will be part of future work. 
\begin{rem}\label{Prop: Sampling}
    If $\Sigma$ has small enough eigenvalues then Alg.~\ref{Alg:HM - Riemannian Random Walk} samples from Eq.~\eqref{Eq: Likelihoog metric}
\end{rem}
For the following DA experiments we will assume that $\Sigma$ has small eigenvalues and so will sample directly using this distribution. See app.~A for sampling results using the aforementioned method.

\section{Data Augmentation Experiments For Classification}

In this section, we explore the ability of the method to enrich data sets to improve classification results.
\subsection{Augmentation Setting} We first test the augmentation method on three reduced data sets extracted from \emph{well-known} databases MNIST and EMNIST. For MNIST, we select 500 samples applying either a balanced split  or a random split ensuring that some classes are far more represented. For EMNIST, we select 500 samples from 10 classes such that they are composed of both lowercase and uppercase characters so that we end up with a small database with strong variability within classes. These data sets are then split such that 80\% is allocated for training (referred to as the \emph{raw data}) and 20\% for validation. For a fair comparison, we use the original test set (\emph{e.g.} $\sim$1000 samples per class for MNIST) to test the classifiers. This ensures statistically meaningful results while assessing the generalisation power on unseen data. We also validate the proposed DA method on the OASIS database which represents a nice example of day-to-day challenges practitioners have to face and is a benchmark database. We use 2D gray scale MR Images (208x176) with a mask notifying brain tissues and are referred to as the \emph{masked T88 images} in \cite{marcus_open_2007}. We refer the reader to their paper for further image preprocessing details. We consider the binary classification problem consisting in trying to detect MRI of patients having been diagnosed with Alzheimer Disease (AD). We split the 416 images into a training set (70\%) (\emph{raw data}), a validation set (10\%) and a test set (20\%).  A summary of demographics, mini-mental state examination (MMSE) and global clinical dementia rating (CDR) is made available in Table.~\ref{table:population}. On the one hand, for each data set, the train set (\emph{raw data}) is augmented by a factor 5, 10 and 15 using classic DA methods (random noise, cropping etc.). On the other hand, VAE models are trained individually on each class of the \emph{raw data}. The generative models are then used to produce 200, 500, 1k or 2k synthetic samples per class with either the classic generation scheme (\emph{i.e.} the prior) or the proposed method. We then train classifiers with 5 independent runs on 1) the \emph{raw data}; 2) the augmented data using basic transformations; 3) the augmented data using the VAE models; 4) only the synthetic data generated by the VAEs. A DenseNet model\footnote{We use the code in \cite{amos_bamosdensenetpytorch_2020} (See App.~E).} \cite{huang_densely_2017} is used for the toy data while we also train hand made MLP and CNN models on OASIS (See App.~E). The main metrics obtained on the test set are reported in Tables.~\ref{Tab: Data Augmentation Toys} and \ref{Tab: Data Augmentation OASIS}. 

  \begin{table}[!t]
\caption{Summary of OASIS database demographics, mini-mental state examination (MMSE) and global clinical dementia rating (CDR) scores.}
\label{table:population}
\begin{center}
\setlength{\tabcolsep}{5pt}
\begin{tabular}{l c c c c c l}
\hline
Data set & Label & Obs. & Age & Sex M/F & MMSE & CDR \\
\hline
\multirow{2}{*}{OASIS} & CN & 316 & $45.1 \pm 23.9$ & 119/197 & $29.1 \pm 1.1$ & 0: 316 \\ 
                       & AD & 100 & $76.8 \pm 7.1$  & 41/59 & $24.3 \pm 4.1$ & 0.5: 70 , 1: 28,  2: 2 \\ 

\hline
\hline
\multirow{2}{*}{Train} & CN & 220 & $45.6 \pm 23.6$ & 86/134 & $29.1 \pm 1.2$ & 0: 220 \\ 
                       & AD & 70  & $77.4 \pm 6.8$  & 29/41  & $23.7 \pm 4.3$ & 0.5: 47 , 1: 21,  2: 2 \\ 
\hline
\multirow{2}{*}{Val}   & CN & 30 & $48.9 \pm 24.1$ & 11/19 & $29.2 \pm 0.8$ & 0: 30 \\ 
                       & AD & 12 & $75.4 \pm 7.2$  & 4/8   & $25.8 \pm 4.2$ & 0.5: 7 , 1: 5,  2: 0 \\
\hline
\multirow{2}{*}{Test}  & CN & 66 & $41.7 \pm 24.3$ & 22/44 & $29.0 \pm 1.0$ & 0: 66 \\ 
                       & AD & 18 & $75.1 \pm 7.5$  & 8/10   & $25.8 \pm 2.7$ & 0.5: 16 , 1: 2,  2: 0 \\
\hline
\end{tabular}
\end{center}
\end{table}

 \subsection{Results}

\subsubsection{Toy Data}As expected generating new samples using the proposed method improves their relevance. The method indeed allows for a quite impressive gain in the model accuracy when synthetic samples are added to the real ones (leftmost column of Table.~\ref{Tab: Data Augmentation Toys}). This is even more striking when looking at the rightmost column where only synthetic samples are used to train the classifier. For instance, when only 200 synthetic samples per class for MNIST are generated with a VAE and used to train the classifier, the classic method fails to produce meaningful samples since a loss of 20 pts in accuracy is observed when compared to the \emph{raw data}. Interestingly, our method seems to avoid such an effect. Even more impressive is the fact that we are able to produce synthetic data sets on which the classifier outperforms greatly the results observed on the \emph{raw data} (3 to 6 pts gain in accuracy) while keeping a relatively low standard deviation (see gray cells). Secondly, this example also shows why geometric DA is still questionable and remains data dependent. For instance, augmenting the \emph{raw data} by a factor 10 (including flips and rotations) does not seem to have a notable effect on the MNIST data sets but still improves results on EMNIST. On the contrary, our method seems quite \textbf{robust to data set changes}.

 \subsubsection{OASIS}Balanced accuracy obtained on OASIS with 3 classifiers is made available in Table.~\ref{Tab: Data Augmentation OASIS}. In this experiment, using the new generation scheme again improves overall the metric for each classifier when compared to the \emph{raw data} and other augmentation methods. Moreover, the strong relevance of the created samples is again supported by the fact that the classifiers are again able to strongly outperform the results on the \emph{raw data} even when trained only with synthetic ones. Finally, the method appears \textbf{robust to classifiers} and can be used with high-dimensional complex data such as MRI.

\begin{table}[t]
    \caption{DA on \emph{toy} data sets. Mean accuracy and standard deviation across 5 independent runs are reported. In gray are the cells where the accuracy is higher on synthetic data than on the \emph{raw data}.}
    \label{Tab: Data Augmentation Toys}
   \centering
   \scriptsize
   \begin{sc}
     \begin{tabular}{c|ccc|ccc}
     \toprule
    Data sets     &  MNIST&  MNIST** &EMNIST** & MNIST &  MNIST**&EMNIST**  \\

       \midrule
       \midrule
        Raw data   &  89.9 (0.6) & 81.6 (0.7) & 82.6 (1.4) &-&-&-  \\
       \midrule
       \multicolumn{4}{c|}{Raw + Synthetic} & \multicolumn{3}{c}{Synthetic only}\\
       \midrule
        Aug. (X5)  &  92.8 (0.4) & 86.5 (0.9) & 85.6 (1.3)  & - & - & - \\
        Aug. (X10) &  88.3 (2.2) & 82.0 (2.4) & 85.8 (0.3)  & - & - & - \\
        Aug. (X15) &  92.8 (0.7) & 85.9 (3.4) & 86.6 (0.8)  & - & - & - \\
       \midrule
       VAE-200*  & 88.5 (0.9) & 84.1 (2.0) & 81.7 (3.0) &  69.9 (1.5) & 64.6 (1.8) & 65.7 (2.6)  \\
       VAE-500*  & 90.4 (1.4) & 87.3 (1.2) & 83.4 (1.6) &  72.3 (4.2) & 69.4 (4.1) & 67.3 (2.4) \\
       VAE-1k*    & 91.2 (1.0) & 86.0 (2.5) & 84.4 (1.6) & 83.4 (2.4) & 74.7 (3.2) & 75.3 (1.4) \\
       VAE-2k*    & 92.2 (1.6) & 88.0 (2.2) & 86.0 (0.2) & 86.6 (2.2) & 79.6 (3.8) & 78.9 (3.0) \\
       \midrule
       \tiny RHVAE-200* & 89.9 (0.5) & 82.3 (0.9) & 83.0 (1.3) &  76.0 (1.8) & 61.5 (2.9) & 59.8 (2.6) \\
       \tiny RHVAE-500* & 90.9 (1.1) & 84.0 (3.2) & 84.4 (1.2) &  80.0 (2.2) & 66.8 (3.3) & 67.0 (4.0) \\
       \tiny RHVAE-1k*  & 91.7 (0.8) & 84.7 (1.8) & 84.7 (2.4) &  82.0 (2.9) & 69.3 (1.8) & 73.7 (4.1) \\
       \tiny RHVAE-2k*  & 92.7 (1.4) & 86.8 (1.0) & 84.9 (2.1) &  85.2 (3.9) & 77.3 (3.2) & 68.6 (2.3) \\
       \midrule
       Ours-200* & 91.0 (1.1)           & 84.1 (2.0)           & 85.1 (1.1)           &  87.2 (1.1)                     & 79.5 (1.6)                & 77.1 (1.6)  \\
       Ours-500* & 92.3 (1.1)           & 87.7 (0.9)           & 85.1 (1.1)           &  89.1 (1.3)                     & 80.4 (2.1)                & 80.2 (2.0) \\
       Ours-1k*  & 93.3 (0.8)           & \textbf{89.7 (0.8)}  & 87.0 (1.0)           &  \cellcolor{gray!30}90.2 (1.4) &\cellcolor{gray!30}86.2 (1.8)&\cellcolor{gray!30}82.6 (1.3)\\
       Ours-2k*  & \textbf{94.3 (0.8)} & 89.1 (1.9)           & \textbf{87.6 (0.8)} &  \cellcolor{gray!30}\textbf{92.6 (1.1)}&\cellcolor{gray!30}\textbf{87.6 (1.3)}&\cellcolor{gray!30}\textbf{86.0 (1.0)} \\
       \bottomrule
       \multicolumn{5}{l}{\tiny{* Number of generated samples} \tiny{** Unbalanced data sets}} \\
     \end{tabular}
     \end{sc}
   \end{table}

\begin{table}[t]
  \caption{DA on OASIS data base. Mean balanced accuracy on independent 5 runs with several classifiers.}
  \label{Tab: Data Augmentation OASIS}
  \centering
\scriptsize
\begin{sc}
  \begin{tabular}{c|cc|cc|cc}
  \toprule
  Networks   & \multicolumn{2}{c}{MLP} & \multicolumn{2}{c}{CNN} & \multicolumn{2}{c}{Densenet} \\
    \midrule
    \midrule
    Raw data            & 80.7 (4.1) &  -  &   72.5 (3.5)   & - & 77.4 (3.3)  & -  \\
    \midrule
             & Raw  +  & Synthetic & Raw  + & Synthetic& Raw  + & Synthetic  \\
             & Synthetic & Only & Synthetic & Only & Synthetic & Only\\
    \midrule
    Aug. (X5)           & 84.3 (1.3) &  -  &   80.0 (3.5)   & - & 73.9 (5.1)  & -   \\
    Aug. (X10)          & 76.0 (2.8) &  -  &   82.8 (3.7)   & - & 78.3 (4.1)  & -    \\
    Aug. (X15)          & 78.7 (5.3) &  -  &   80.3 (3.7)   & - & 76.6 (1.1)  & -    \\
    \midrule
    VAE-200$^{*}  $     & 80.7 (1.5) &  77.8 (1.3)  &  79.4 (3.6)     & 65.0 (12.3)  & 76.5 (3.2)  & 74.0 (3.0)  \\
    VAE-500$^{*}  $     & 79.7 (1.4) &  77.4 (1.5)  &  72.6 (7.0)     & 70.2 (5.0)   & 74.9 (4.3)  & 72.8 (1.8) \\
    VAE-1000$^{*} $     & 81.3 (0.0) &  76.5 (0.6)  &  74.4 (9.4)     & \cellcolor{gray!30}73.0 (3.3)   & 73.5 (1.3)  & 74.9 (2.6) \\
    VAE-2000$^{*} $     & 80.7 (0.3) &  78.1 (1.6)  &  71.1 (4.9)     & \cellcolor{gray!30}76.9 (2.6)   & 74.0 (4.9)  & 73.3 (3.4) \\
    \midrule
    Ours-200$^{*} $     & 84.3 (0.0)            &  \cellcolor{gray!30}86.7 (0.4)  & 76.4 (5.0)           &   \cellcolor{gray!30}75.4 (6.6)     &  78.2 (3.0)          & 74.3 (4.8) \\
    Ours-500$^{*} $     & \textbf{87.2 (1.2)}   &  \cellcolor{gray!30}\textbf{88.6 (1.1)}  & 81.8 (4.6)           &   \cellcolor{gray!30}81.8 (3.7)     &  80.2 (2.8)          & \cellcolor{gray!30}\textbf{84.2 (2.8)} \\
    Ours-1000$^{*}$     & 84.2 (0.3)            &  \cellcolor{gray!30}84.4 (1.8)  & 83.5 (3.2)           &   \cellcolor{gray!30}79.8 (2.8)     &  82.2 (4.7)          & 76.7 (3.8) \\
    Ours-2000$^{*}$     & 85.3 (1.9)            &  \cellcolor{gray!30}84.2 (3.3)  & \textbf{84.5 (1.9)}  &   \cellcolor{gray!30}\textbf{83.9 (1.9)}     &  \textbf{82.9 (1.8)} & 73.6 (5.8) \\

    \bottomrule
    \multicolumn{5}{l}{\tiny{* Number of generated samples}} 
  \end{tabular}
\end{sc}
\end{table}

\section{Conclusion}
In this paper, we proposed a new way to generate new data from a Variational Autoencoder which has learned the latent geometry of the input data. This method was then used to perform DA to improve classification tasks in the low sample size setting on both toy and real data and with different kind of classifiers. In each case, the method allows for an impressive gain in the classification metrics (\emph{e.g.} balanced accuracy jumps from 80.7 to 88.6 on OASIS). Moreover, the relevance of the generated data was supported by the fact that classifiers were able to perform better when trained with only synthetic data than on the \emph{raw data} in all cases. Future work would consist in using the method on even more challenging data such as 3D volumes and using smaller data sets.

\section*{Acknowledgment}

The research leading to these results has received funding from the French government under management of Agence Nationale de la Recherche as part of the ``Investissements d'avenir'' program, reference ANR-19-P3IA-0001 (PRAIRIE 3IA Institute) and reference ANR-10-IAIHU-06 (Agence Nationale de la Recher\-che-10-IA Institut Hospitalo-Universitaire-6). Data were provided in part by OASIS: Cross-Sectional: Principal Investigators: D. Marcus, R, Buckner, J, Csernansky J. Morris; P50 AG05681, P01 AG03991, P01 AG026276, R01 AG021910, P20 MH071616, U24 RR021382

\clearpage
\bibliographystyle{splncs04}
\bibliography{references}

\clearpage
\appendix
\section*{Appendix A: Comparison with Prior-Based Methods}

In this section, we compare the samples quality between \emph{prior-based} methods and ours on various standard and \emph{real-life} data sets.

\begin{figure*}[ht]
\adjustbox{minipage=3.em,raise=\dimexpr -3\height}{\small Latent\\ space}
\captionsetup[subfigure]{position=above, labelformat = empty}
\subfloat[\hspace{0.15in}VAE - $\mathcal{N}(0, I_d)$]{\includegraphics[width=1.in]{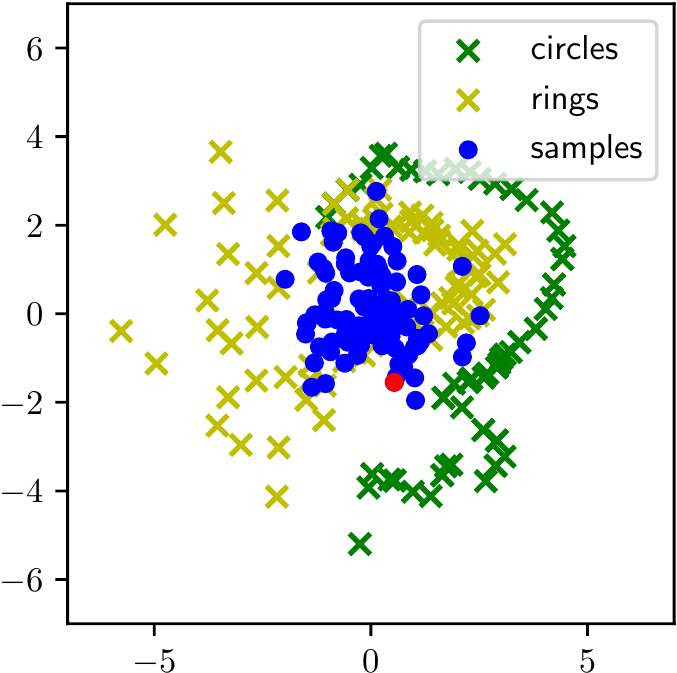}}
\hfil
\hspace{0.01in}
\subfloat[\hspace{0.15in}VAMP - VAE]{\includegraphics[width=1.in]{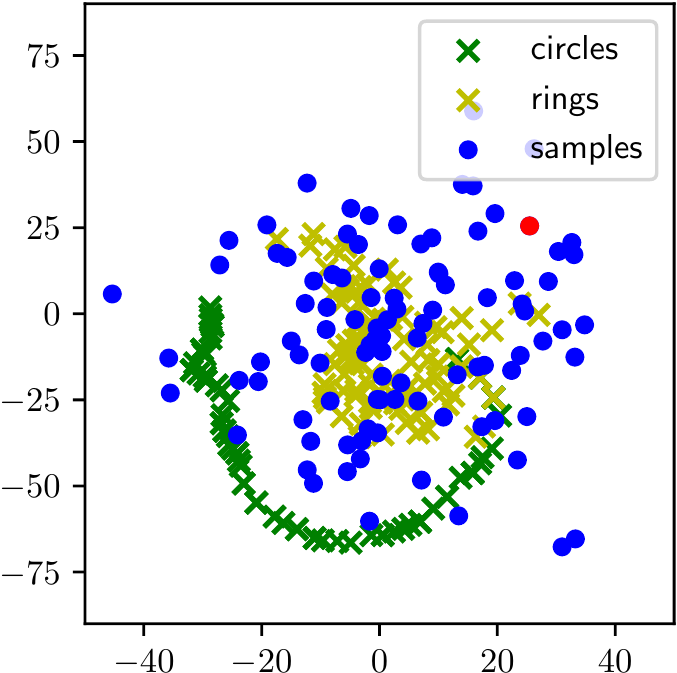}}
\hfil
\subfloat[\hspace{0.15in}RHVAE]{\includegraphics[width=1.1in]{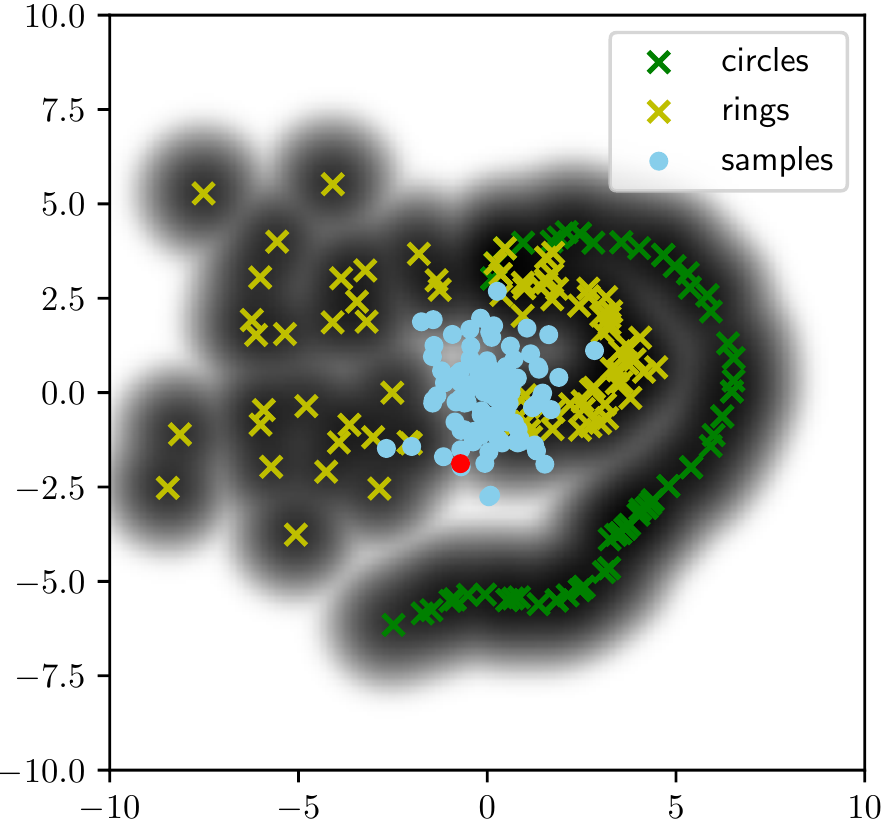}}
\hfil
\subfloat[\hspace{0.15in}Ours]{\includegraphics[width=1.1in]{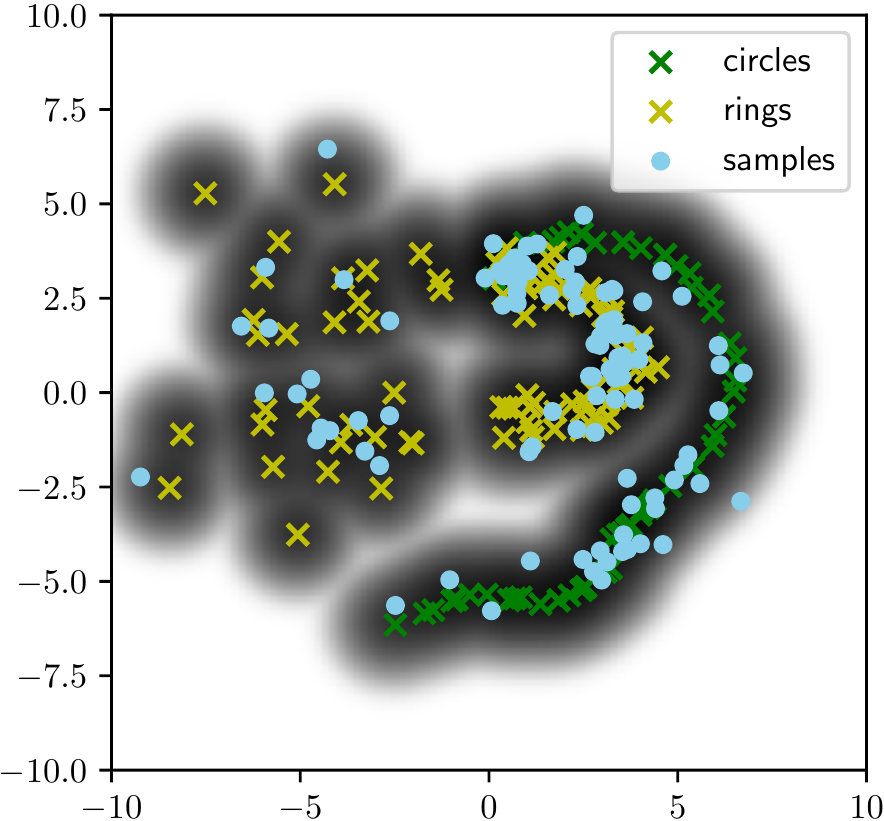}}
\hfil
\vfil
\vspace{-0.1in}
\adjustbox{minipage=3.3em,raise=\dimexpr -3.5\height}{\small Decoded\\ samples}
\subfloat{\includegraphics[width=1.in]{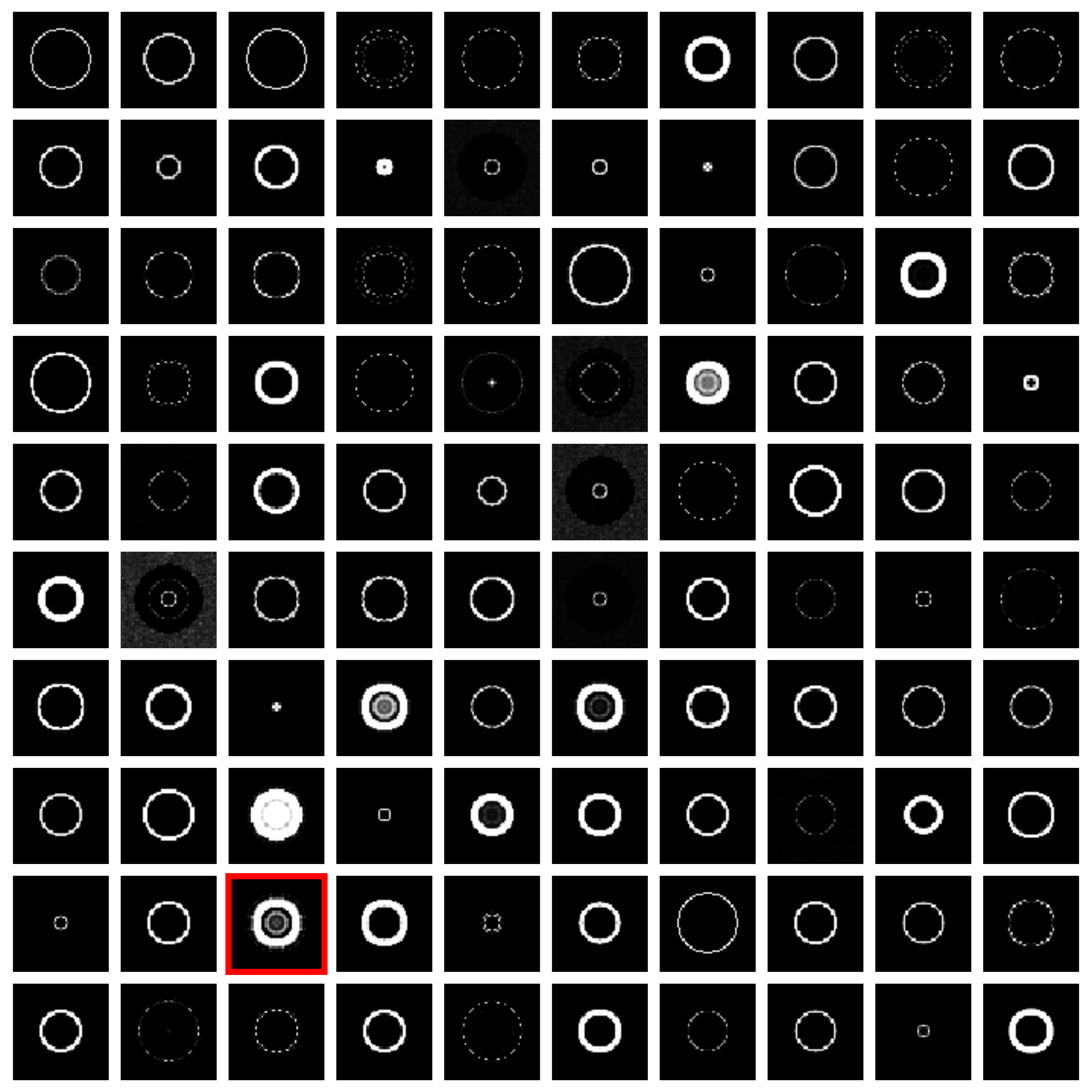}}
\hfil
\subfloat{\includegraphics[width=1.in]{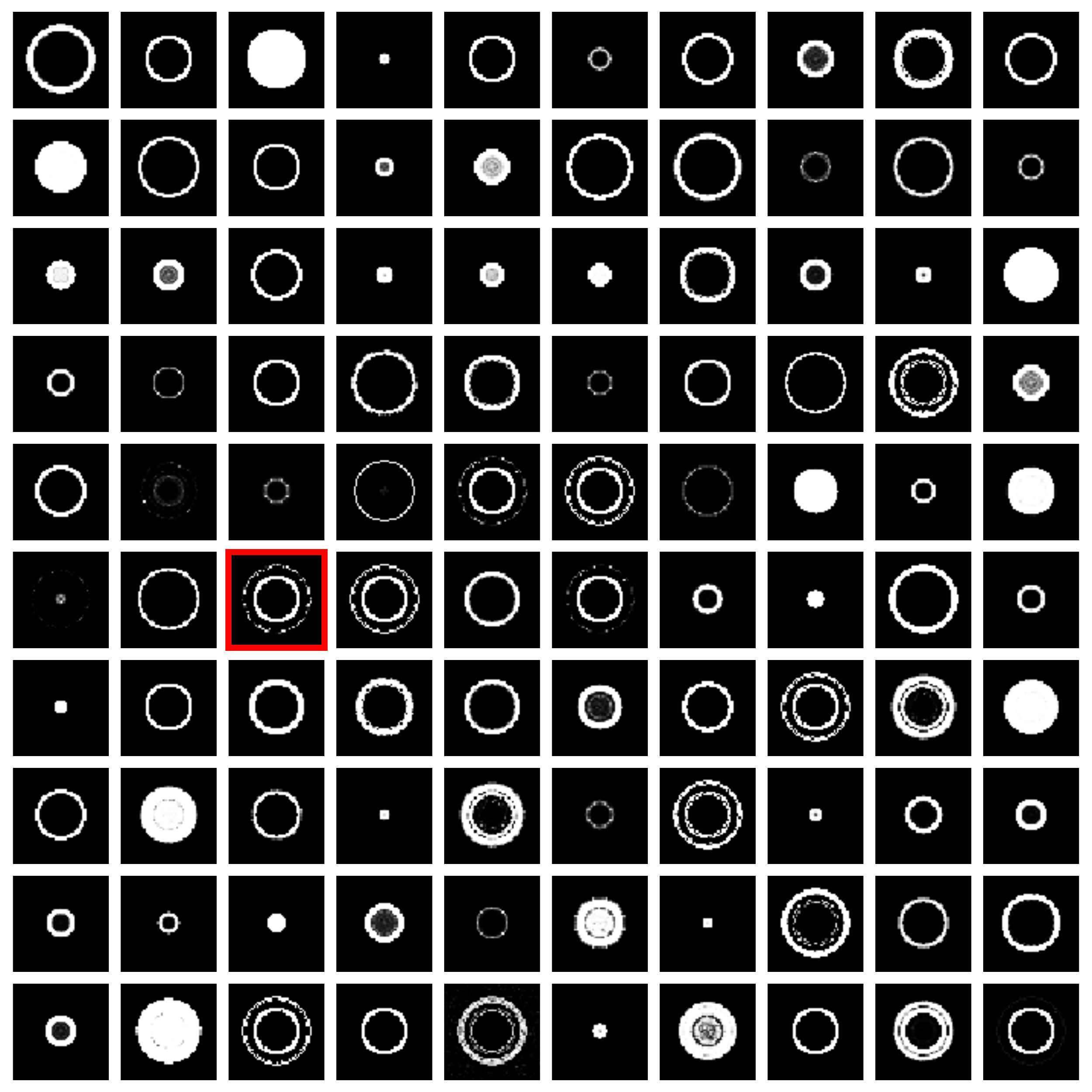}}
\hfil
\subfloat{\includegraphics[width=1.in]{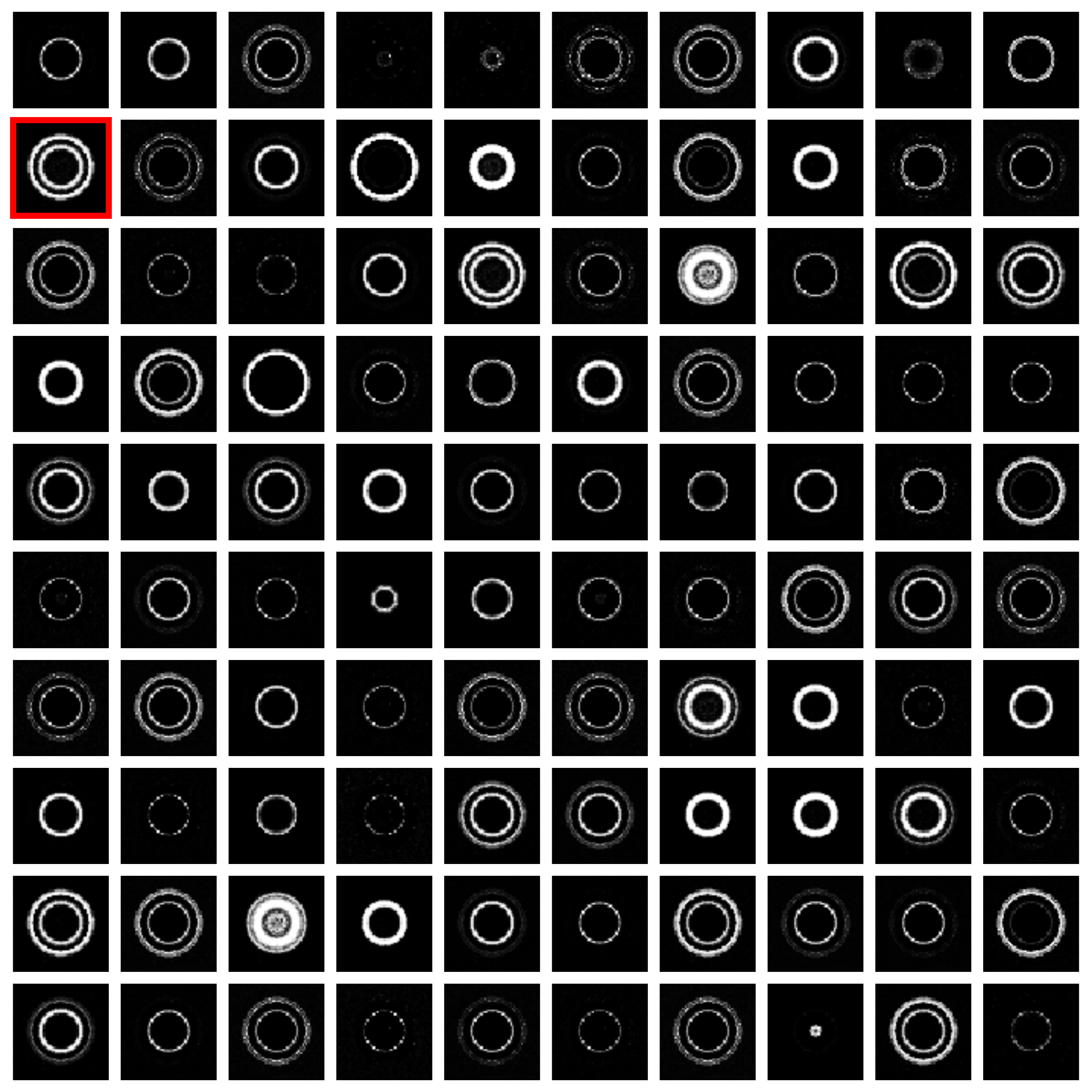}}
\hfil
\subfloat{\includegraphics[width=1.in]{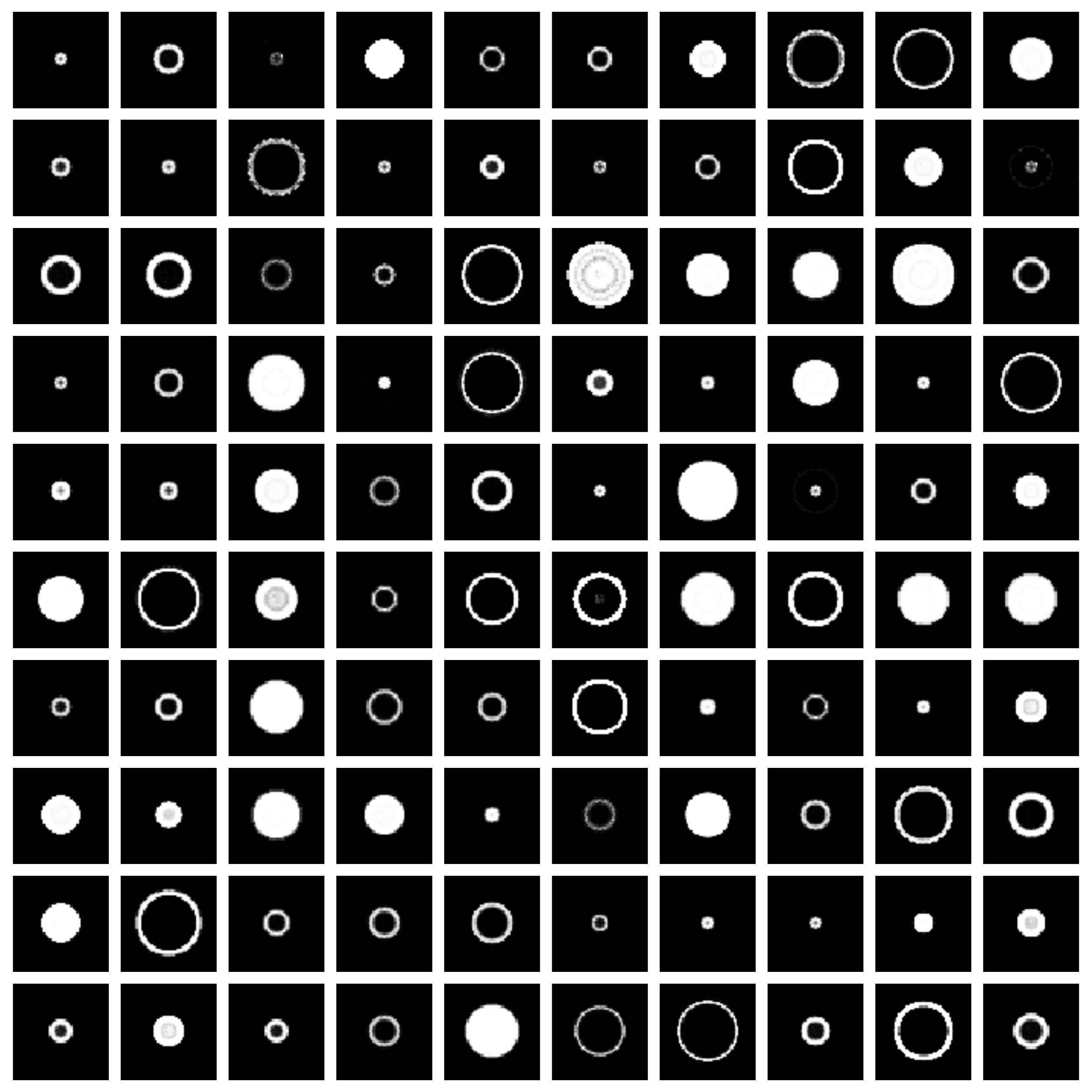}}
\caption{Comparison between prior-based generation methods and the proposed Riemannian random walk (ours). Top: the learned latent space with the encoded training data (crosses) and 100 samples for each method (blue dots). Bottom: the resulting decoded images. The models are trained on 180 binary circles and rings with the same neural network architectures.}
\label{Fig: Shapes Sampling Comparision}
\end{figure*}
\subsubsection{Standard Data Sets}

\begin{figure*}[p]
    \centering
    \adjustbox{minipage=5.5em,raise=\dimexpr -4\height}{\small Training\\ samples}
    \captionsetup[subfigure]{position=above, labelformat = empty}
    \subfloat[MNIST]{\includegraphics[width=1.25in]{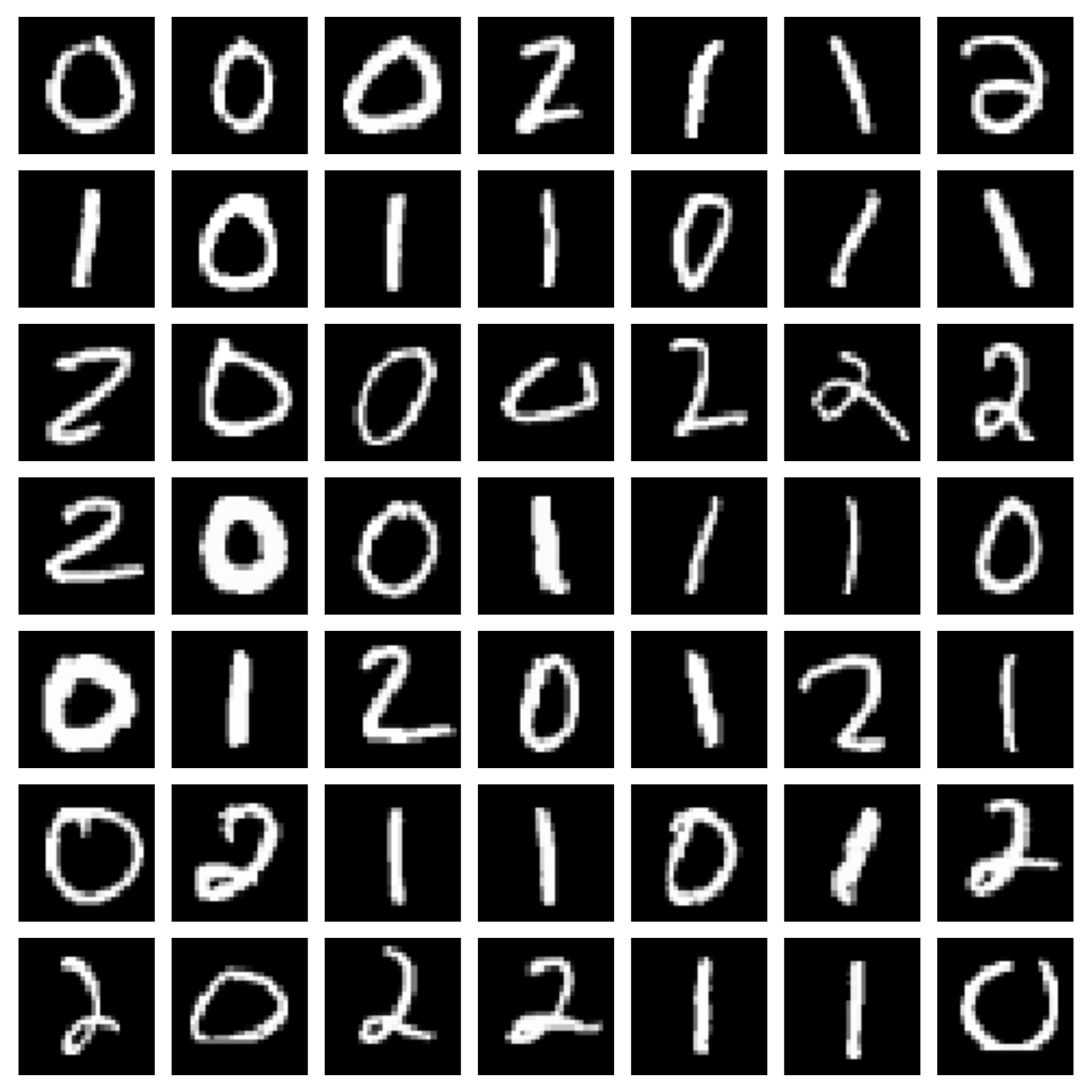}}
    \hspace{0.01in}
    \subfloat[FashionMNIST]{\includegraphics[width=1.25in]{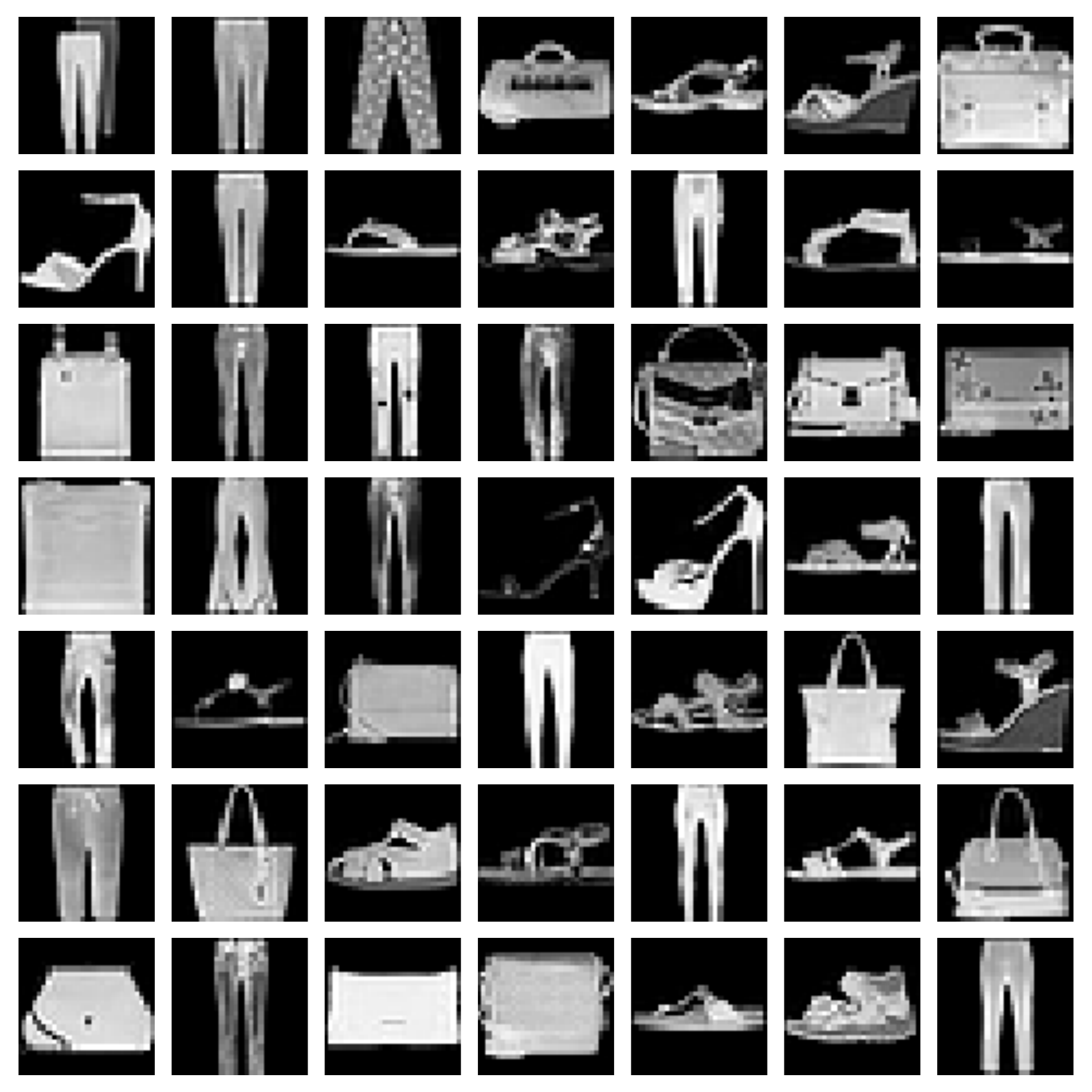}}
    \hspace{0.01in}
    \subfloat[Synthetic data]{\includegraphics[width=1.25in]{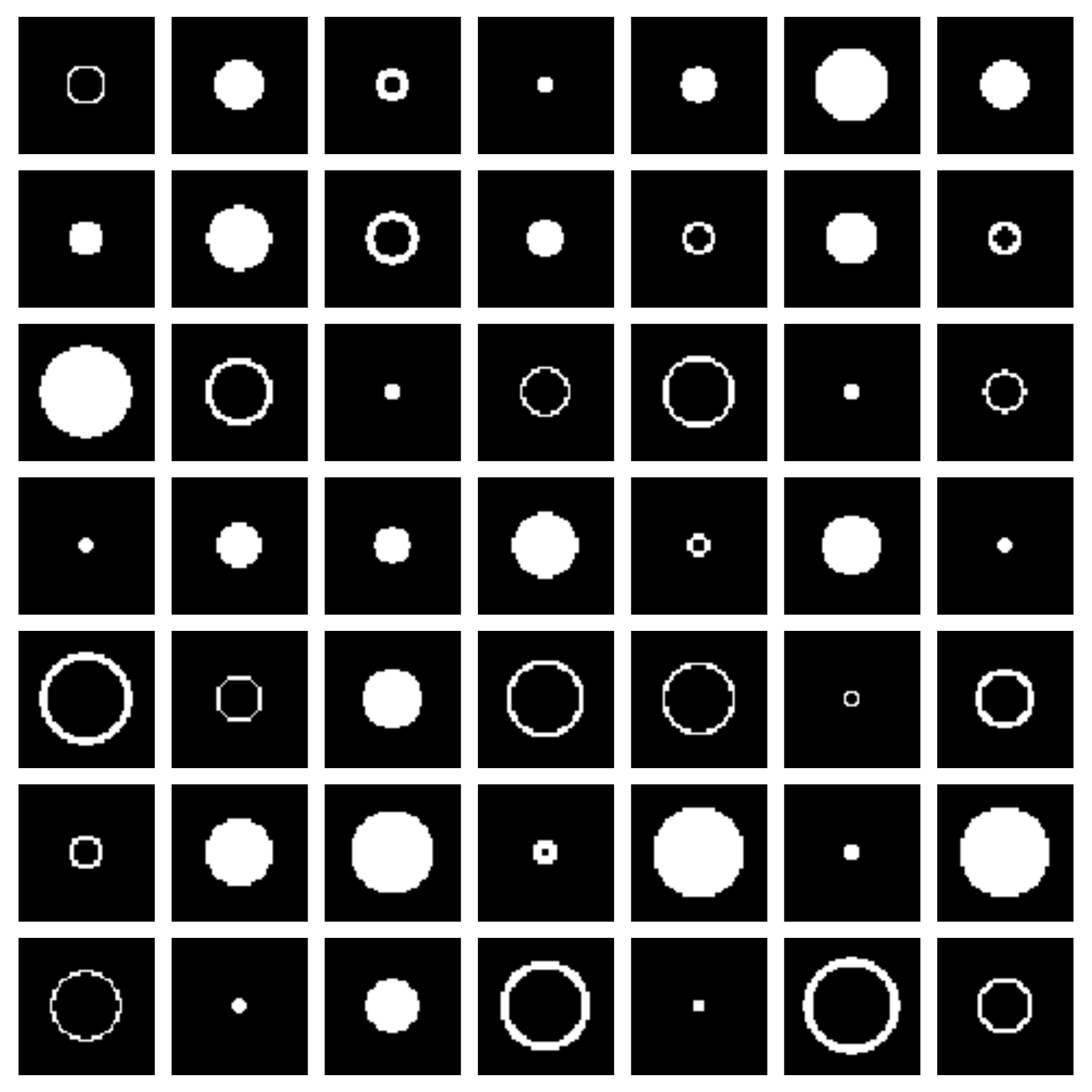}}
    \vspace{-1em}
    \adjustbox{minipage=5.5em,raise=\dimexpr -4\height}{\small VAE +\\ $\mathcal{N}(0, I_d)$}
    \subfloat{\includegraphics[width=1.25in]{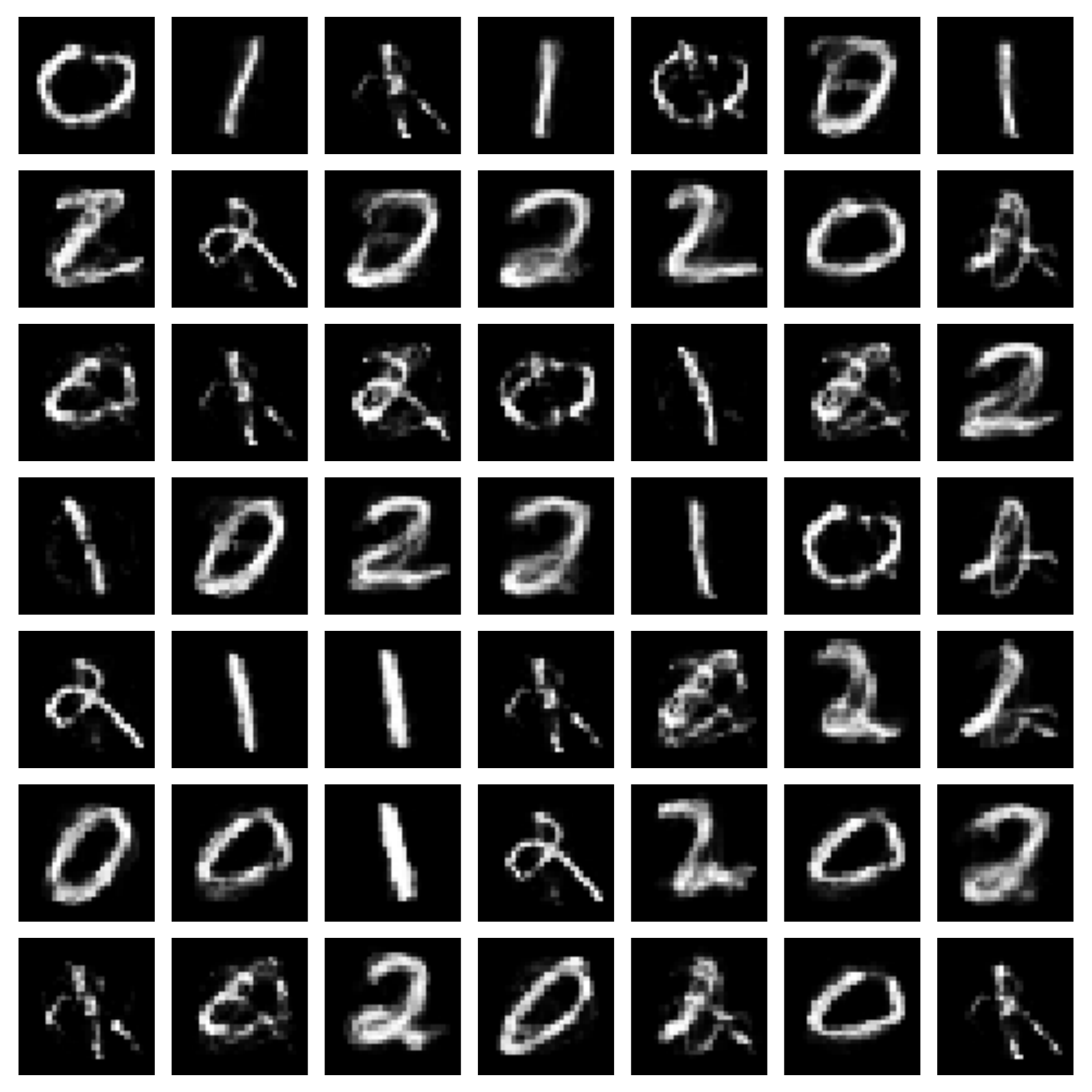}}
    \hspace{0.01in}
    \subfloat{\includegraphics[width=1.25in]{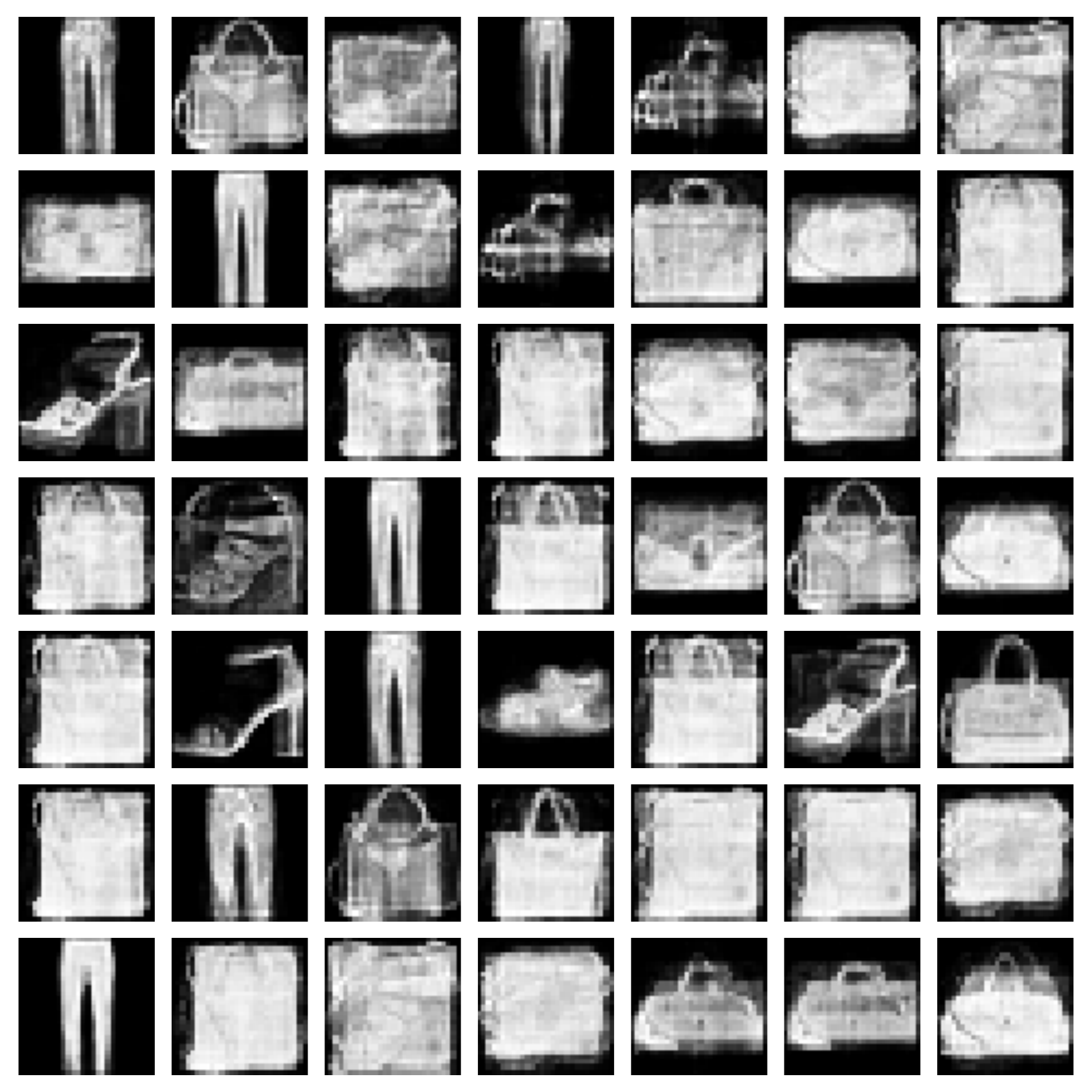}}
    \hspace{0.01in}
    \subfloat{\includegraphics[width=1.25in]{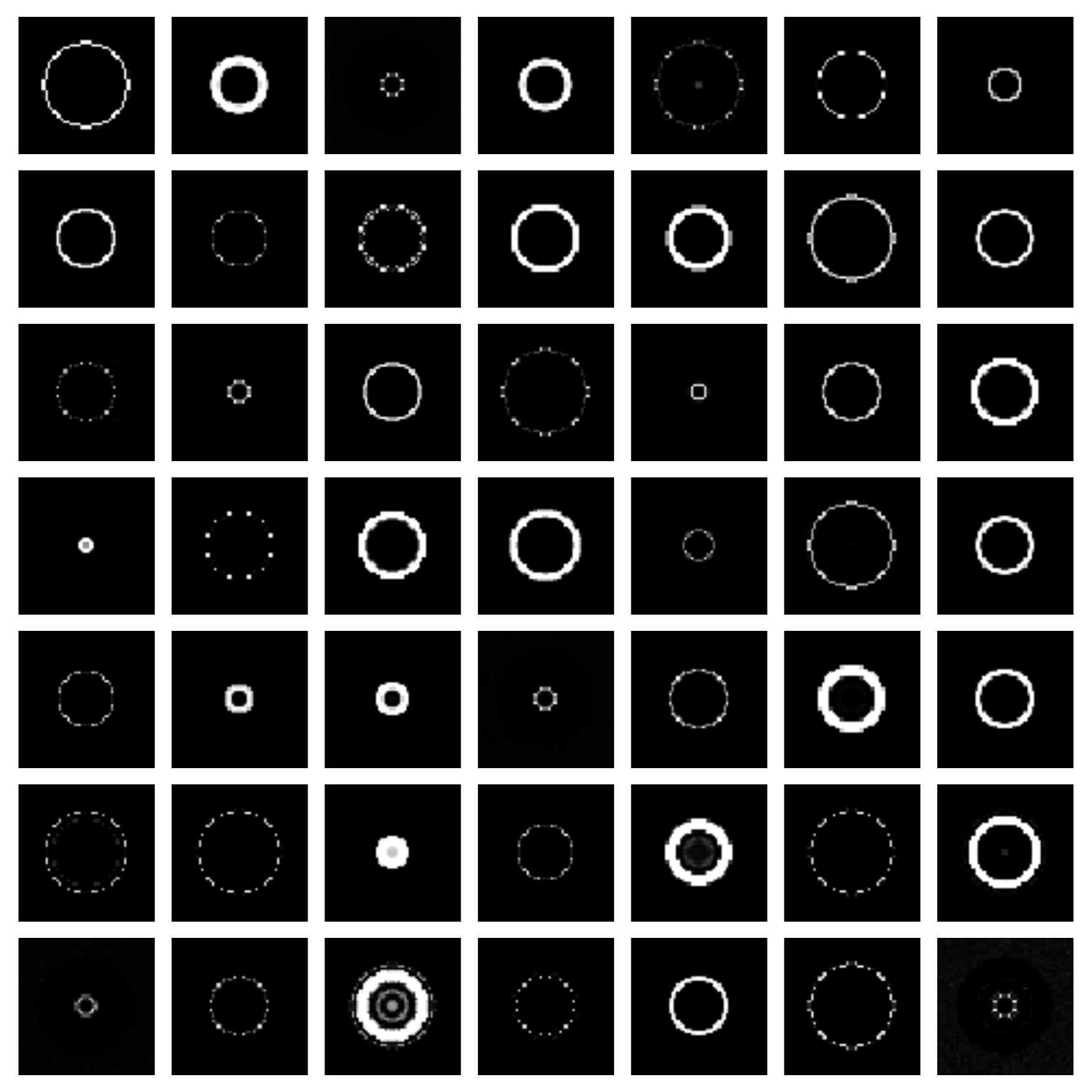}}
    \vspace{-1em}
    \adjustbox{minipage=5.5em,raise=\dimexpr -4\height}{\small VAE +\\ VAMP prior}
    \subfloat{\includegraphics[width=1.25in]{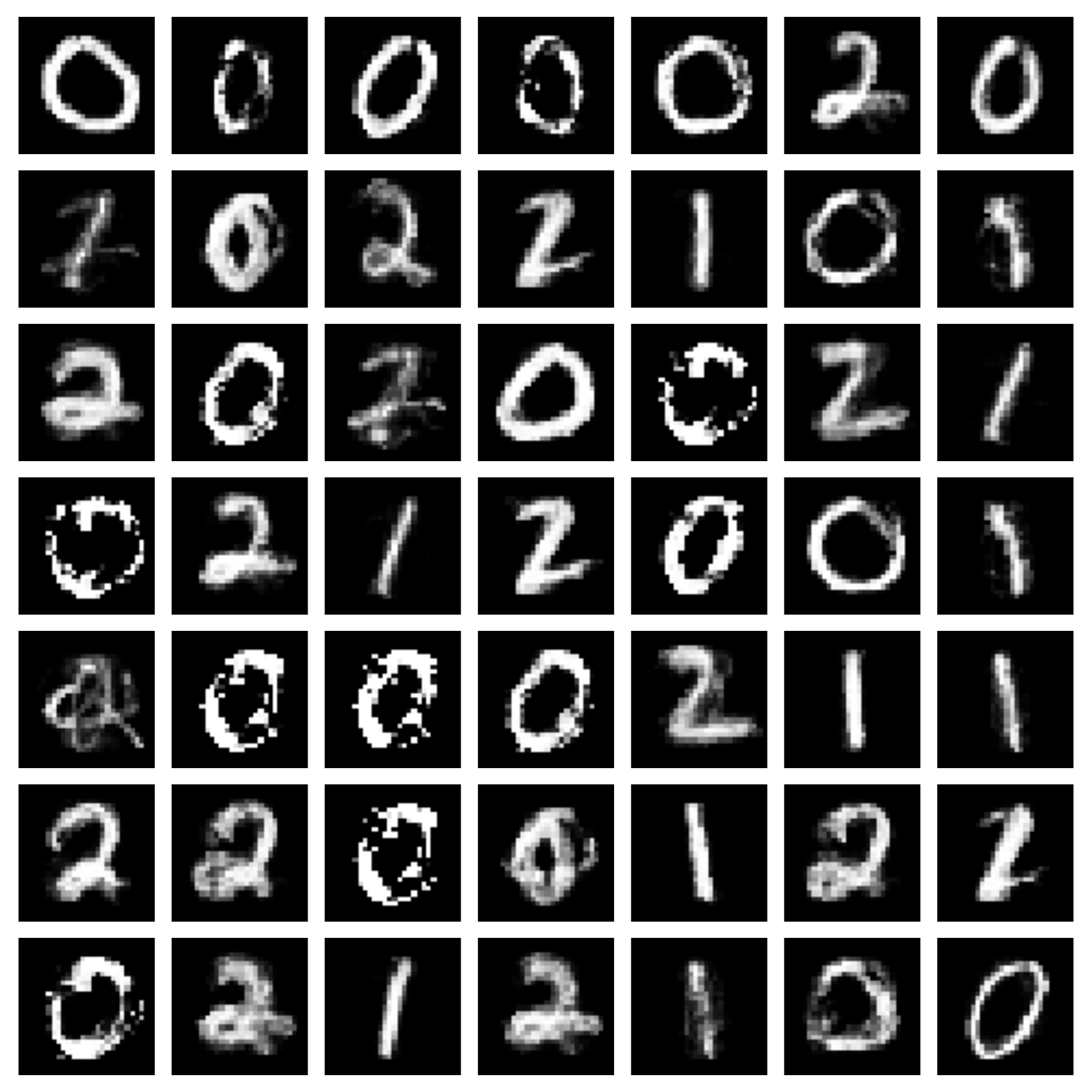}}
    \hspace{0.01in}
    \subfloat{\includegraphics[width=1.25in]{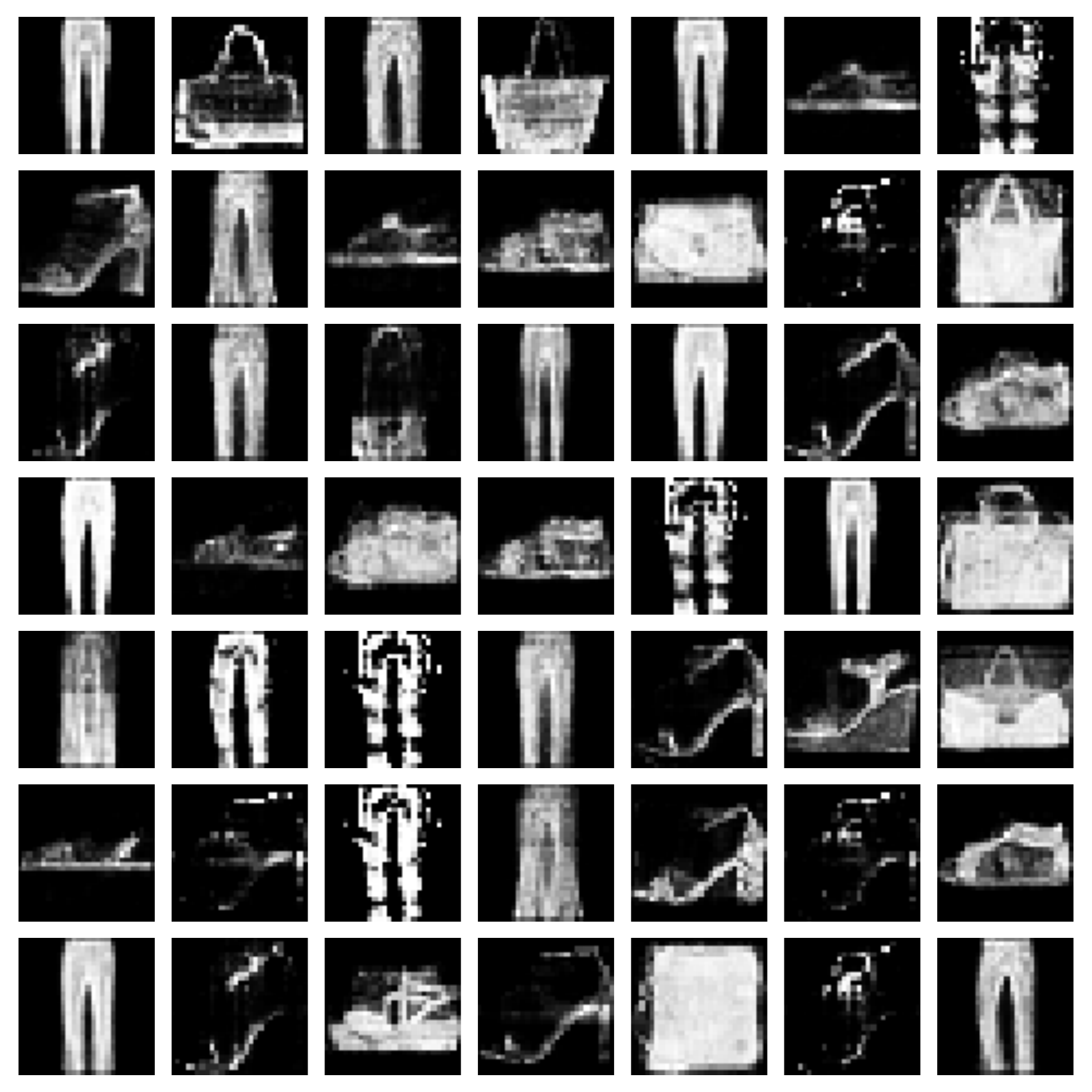}}
    \hspace{0.01in}
    \subfloat{\includegraphics[width=1.25in]{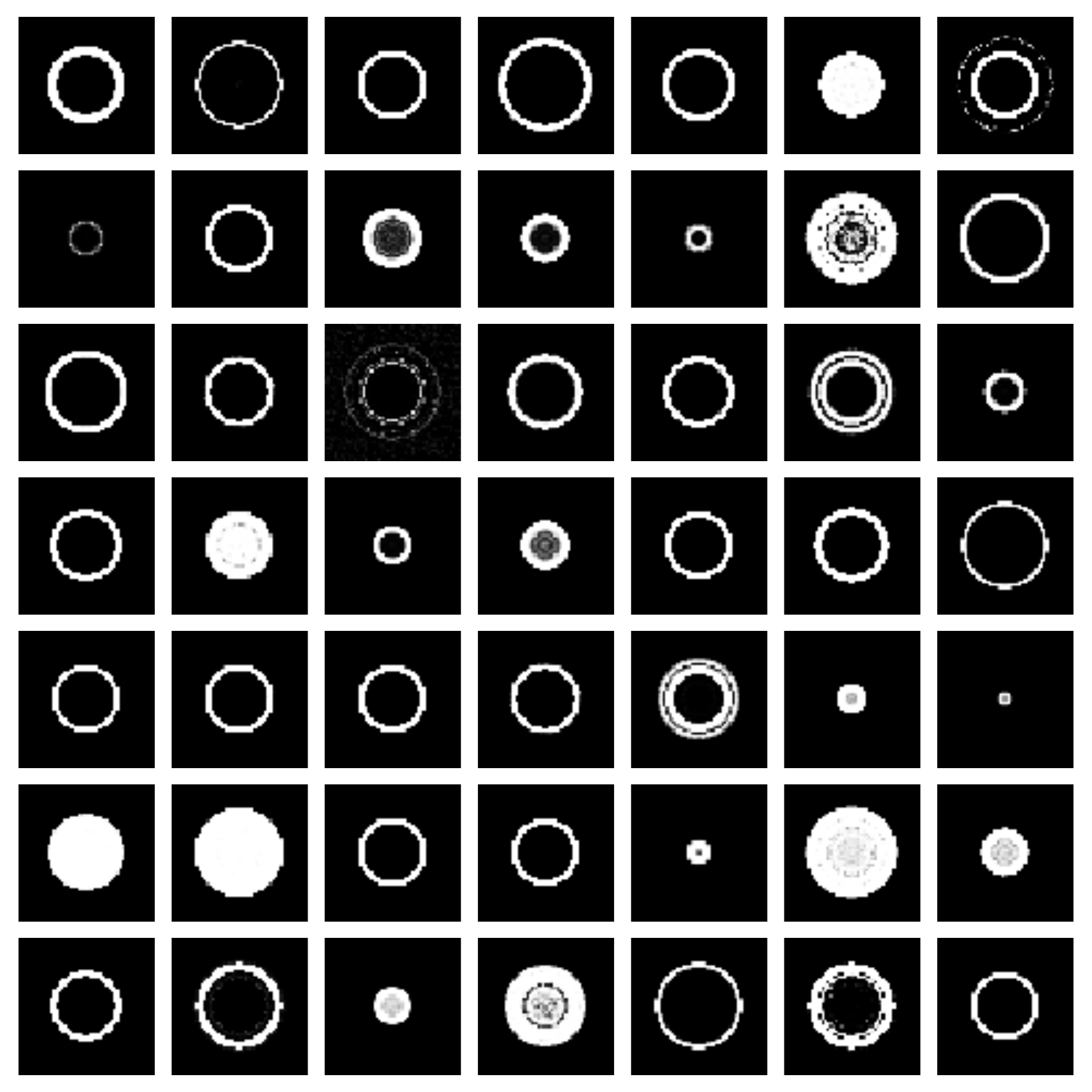}}
    \vspace{-1em}
    \adjustbox{minipage=5.5em,raise=\dimexpr -4\height}{\small RHVAE +\\ $\mathcal{N}(0, I_d)$}
    \subfloat{\includegraphics[width=1.25in]{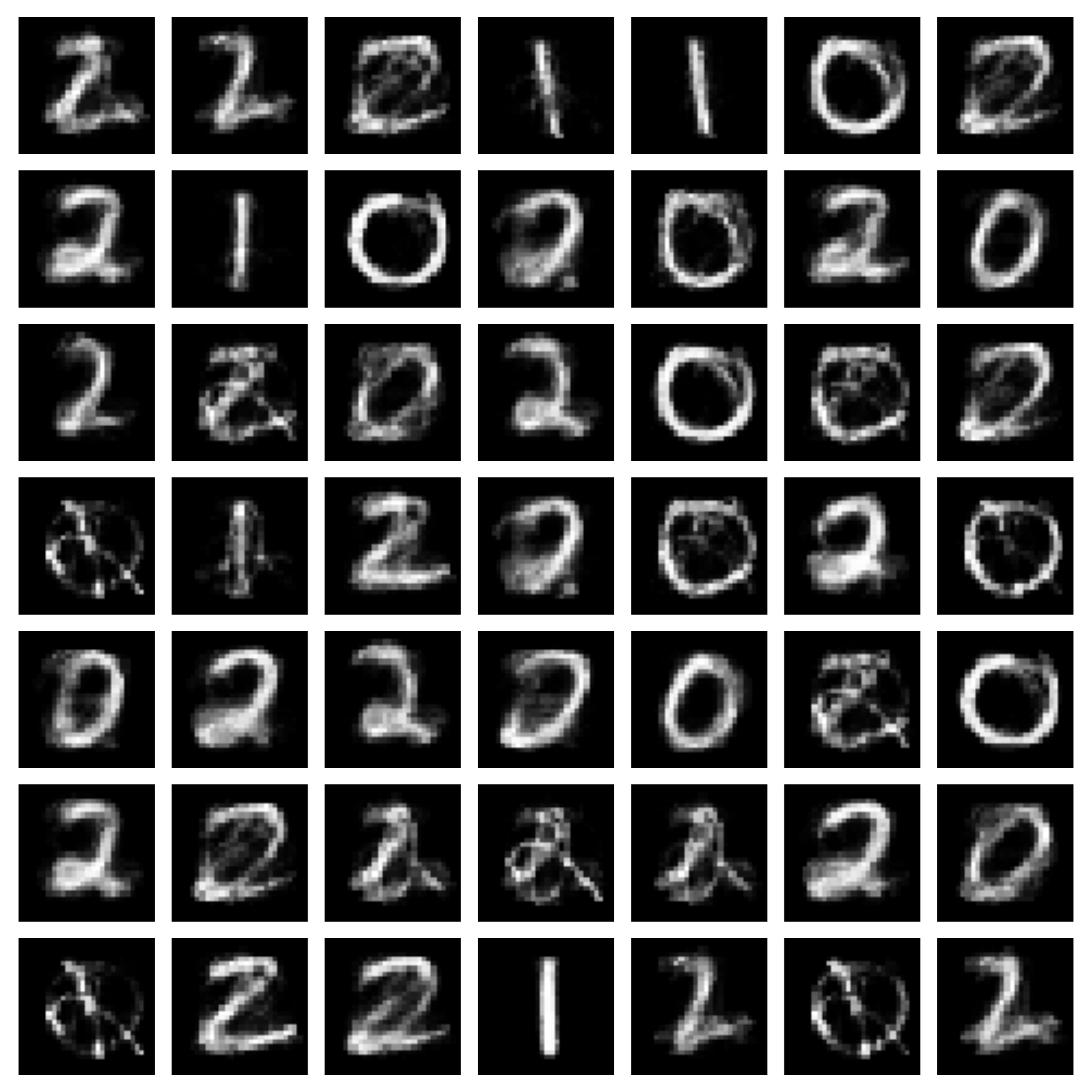}}
    \hspace{0.01in}
    \subfloat{\includegraphics[width=1.25in]{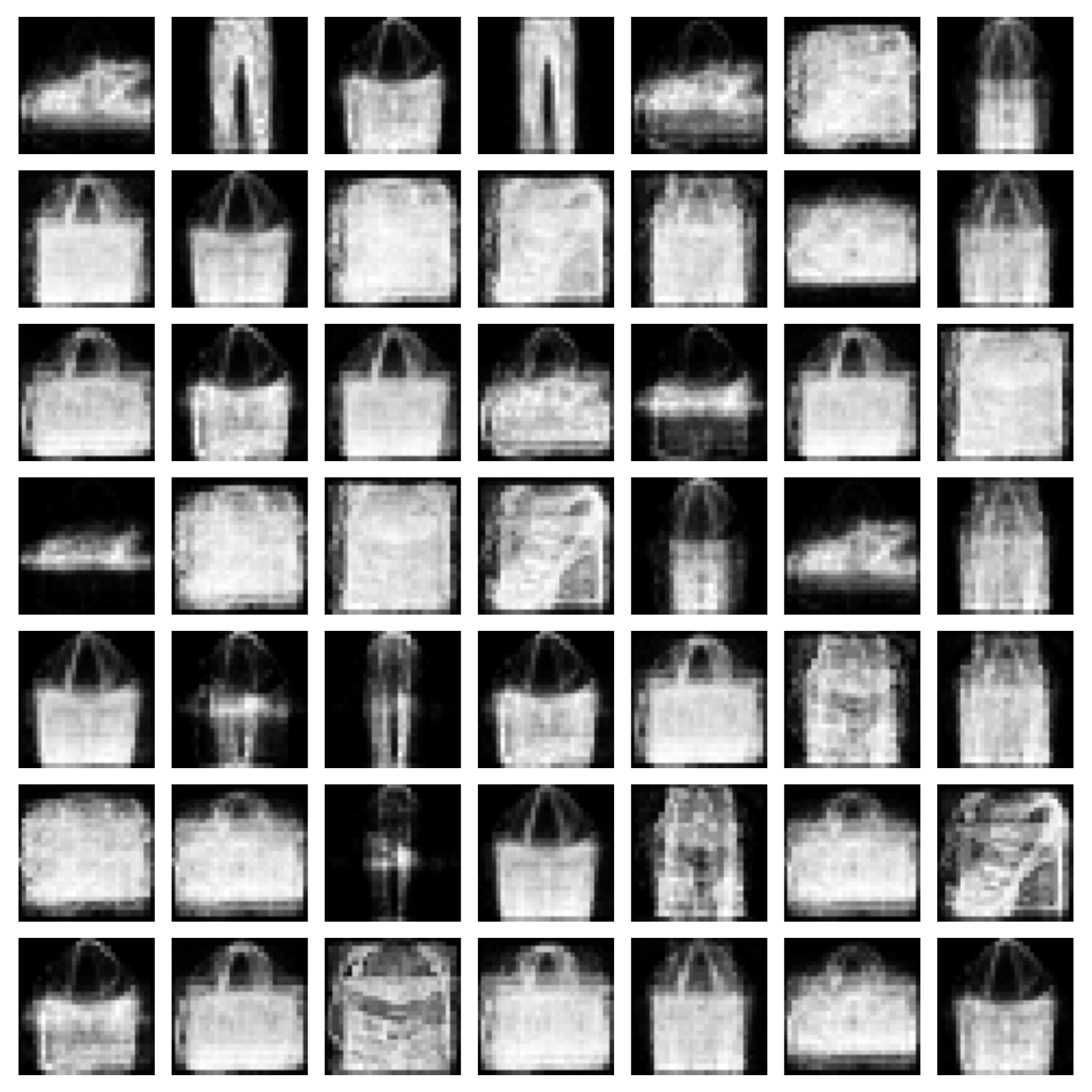}}
    \hspace{-0.01in}
    \subfloat{\includegraphics[width=1.25in]{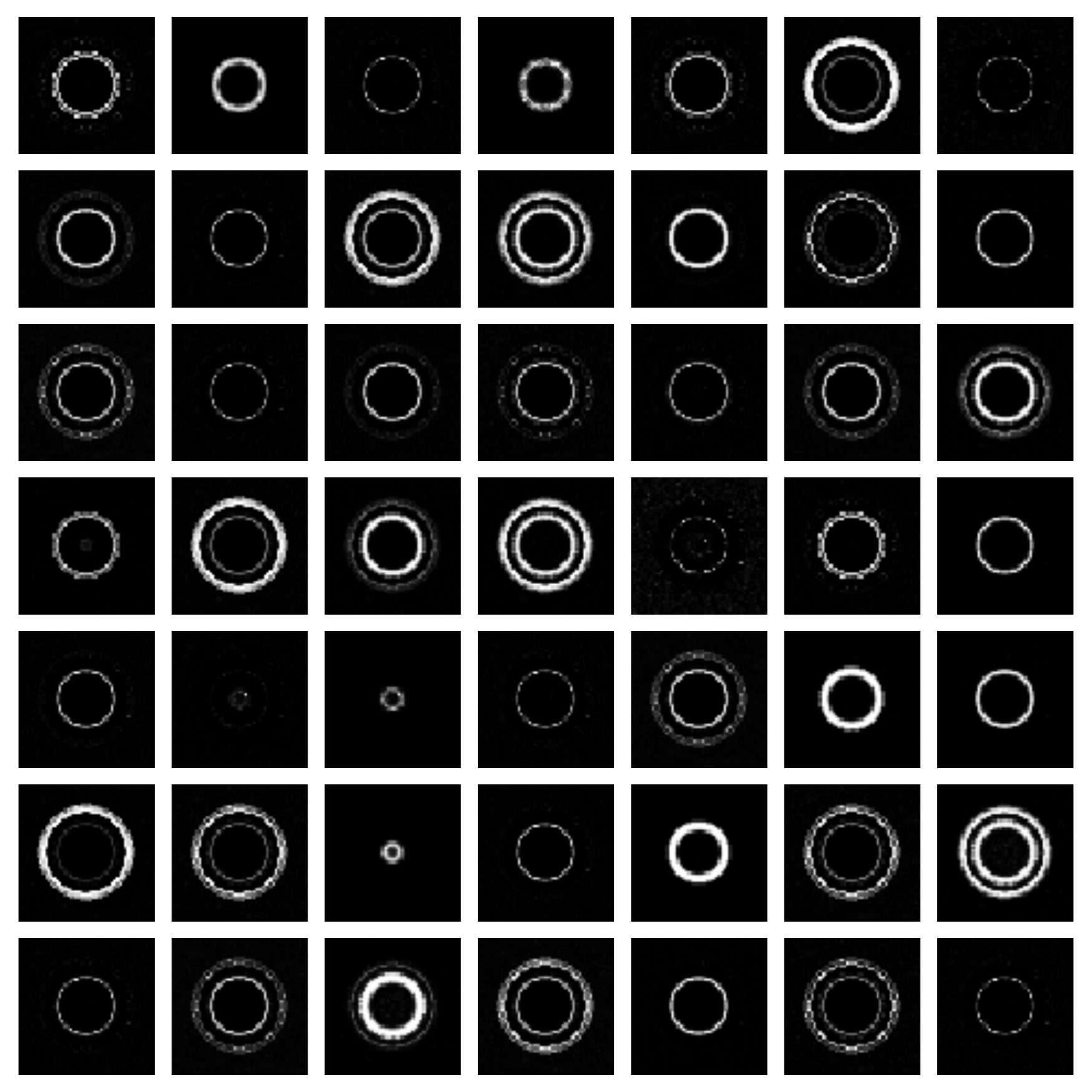}}
    \vspace{-1em}
    \adjustbox{minipage=5.5em,raise=\dimexpr -2\height}{\small RHVAE +\\ Riemannian\\ random walk (Ours)}
    \subfloat{\includegraphics[width=1.25in]{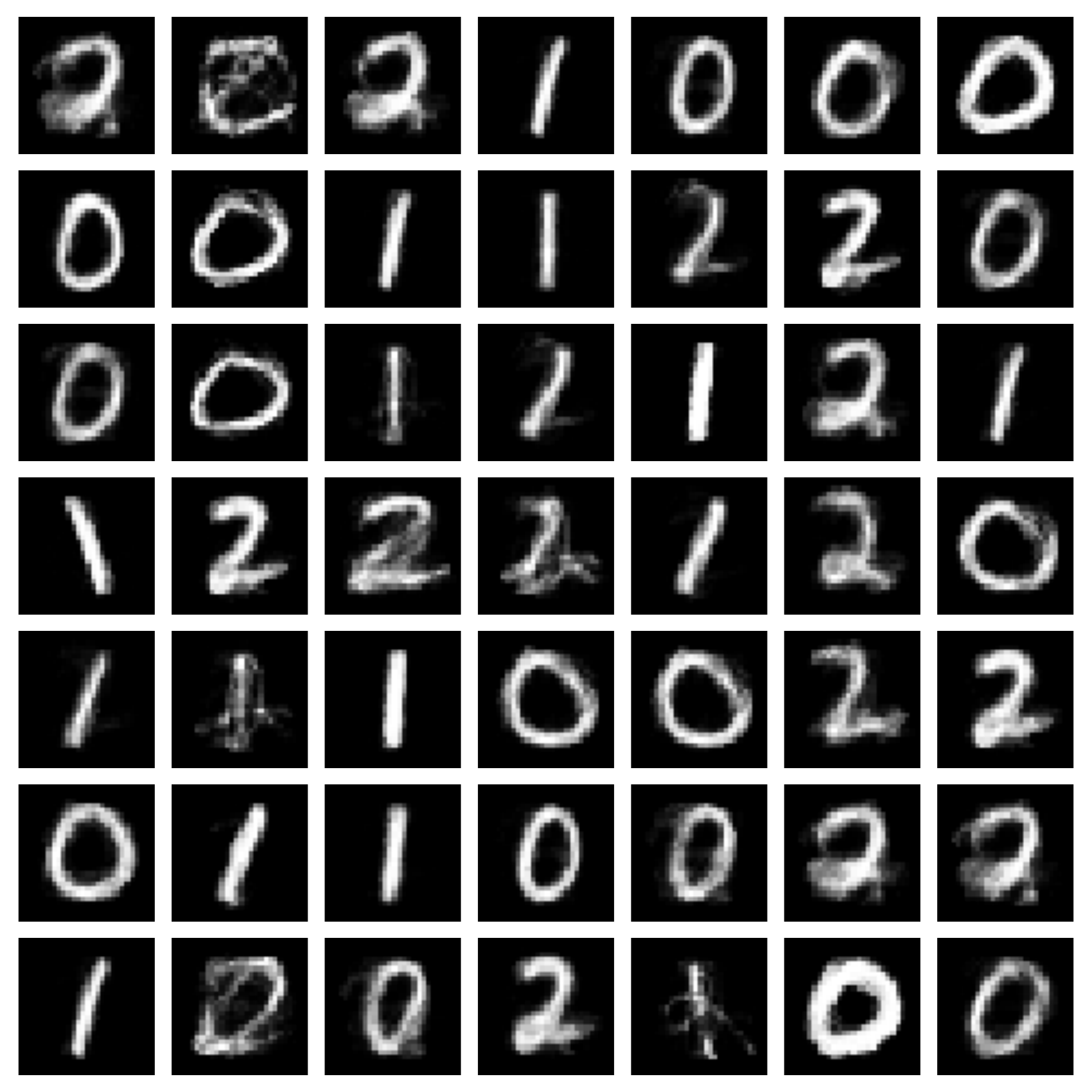}}
    \hspace{0.01in}
    \subfloat{\includegraphics[width=1.25in]{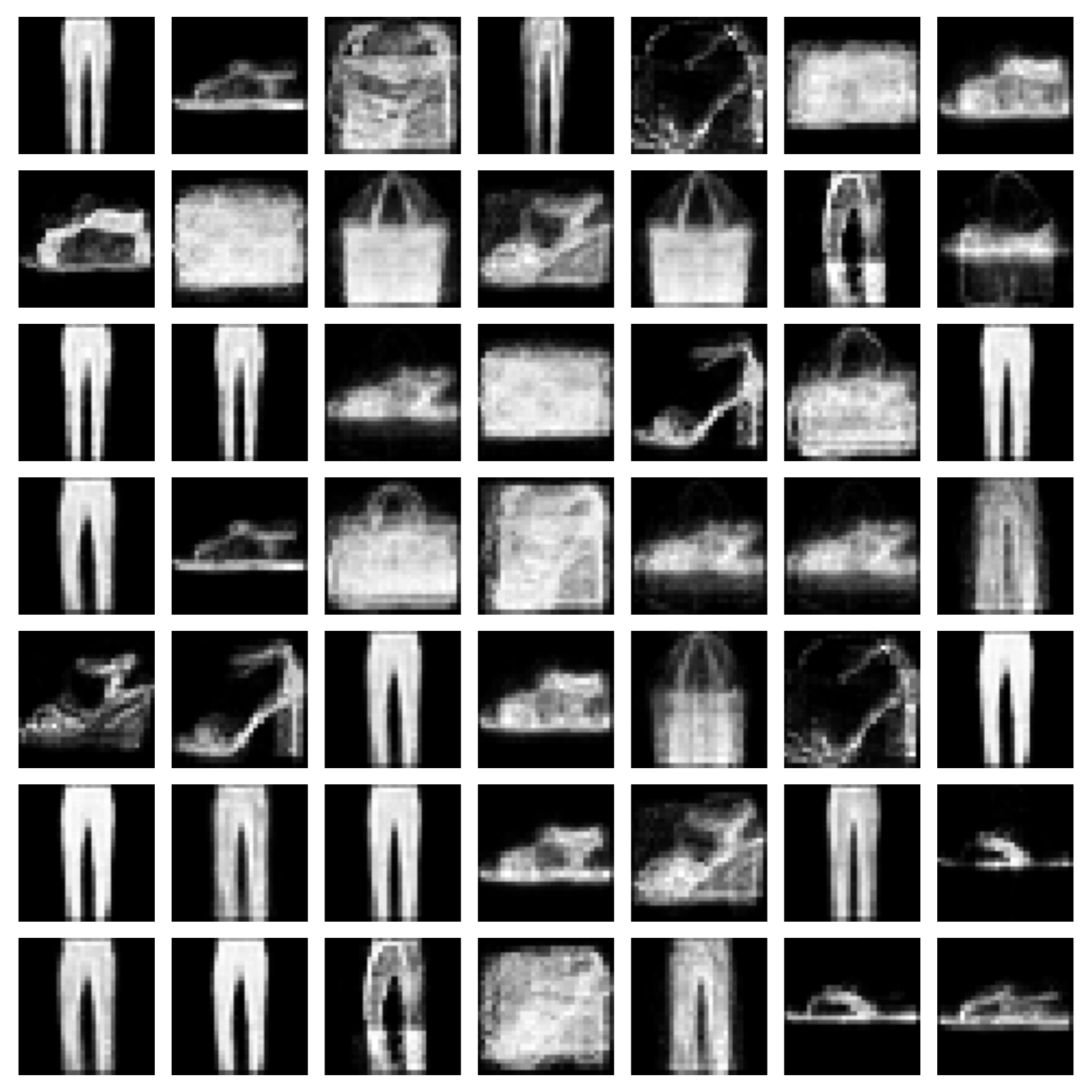}}
    \hspace{0.01in}
    \subfloat{\includegraphics[width=1.25in]{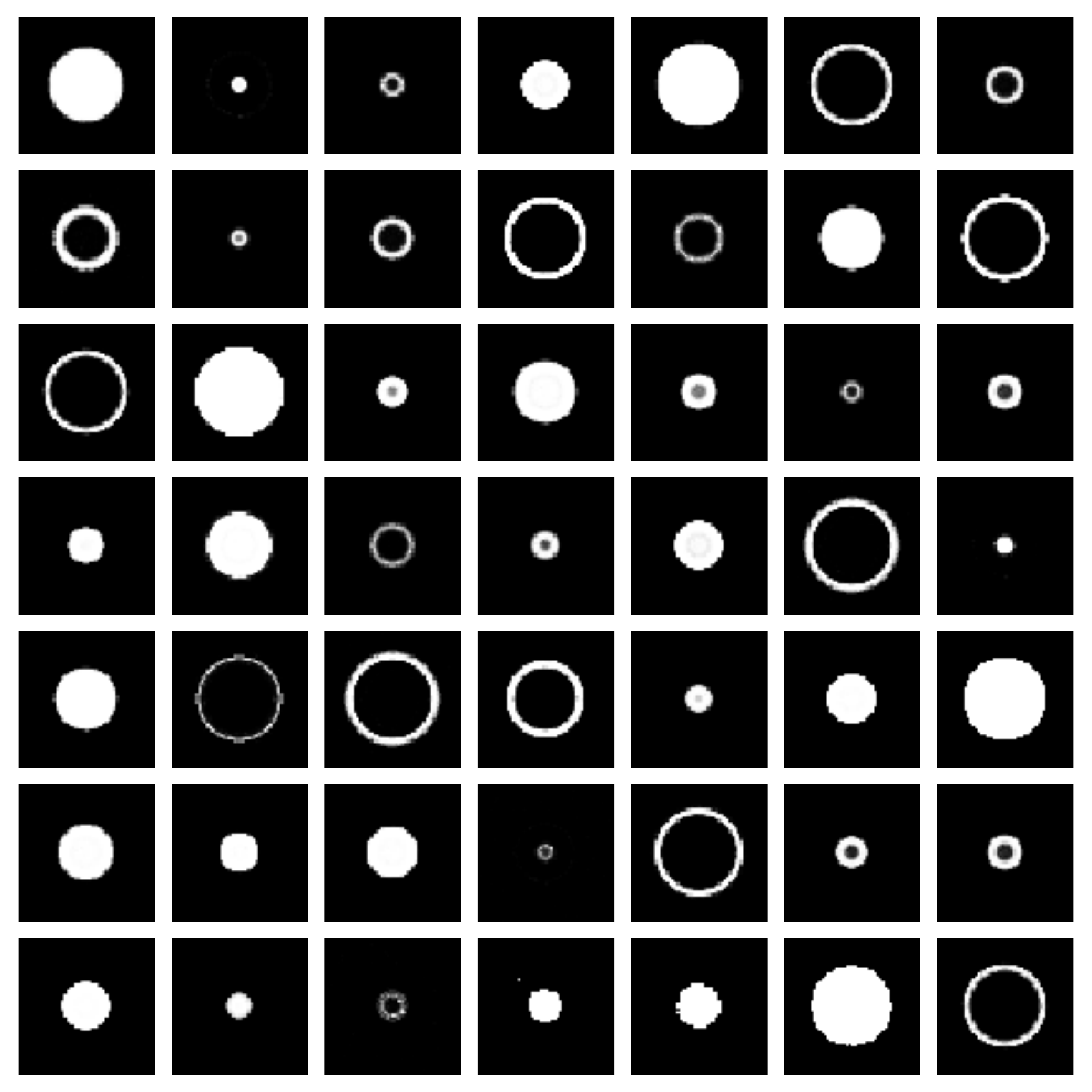}}
    \caption{Comparison of 4 sampling methods on the \emph{reduced} MNIST, \emph{reduced} Fashion and the synthetic data sets. From top to bottom: 1) samples extracted from the training set; 2) samples generated with a Vanilla VAE and using the prior; 3) from the VAMP prior VAE ; 4) from a RHVAE and the \emph{prior-based} generation scheme; 5) from a RHVAE and using the proposed Riemannian random walk. All the models are trained with the same encoder and decoder networks and identical latent space dimension. An early stopping strategy is adopted and consists in stopping training if the ELBO does not improve for 50 epochs.}
    \label{Fig: Comparison}
\end{figure*}

The method is first validated on a hand-made synthetic data set composed of 180 binary images of circles and rings of different diameters and thicknesses (see training samples in Fig.~\ref{Fig: Comparison}). We then train a VAE with a normal prior, a VAE with a VAMP prior \cite{tomczak_vae_2018} and a RHVAE until the ELBO does not improve for 50 epochs. Any relevant parameters setting is stated in App. D.

Fig.~\ref{Fig: Shapes Sampling Comparision} highlights the obtained samplings with each model using either the \emph{prior-based} generation procedure or the one proposed in this paper. The first row presents the learned latent space along with the means of the posteriors associated to the training data (crosses) and 100 latent space samples for each generation method (blue dots). The second row displays the corresponding decoded images. The first outcome of such a study is that sampling from the prior distribution $\mathcal{N}(0, I_d)$ leads to a poor latent space prospecting. Therefore, even with balanced training classes, we end up with a model over-representing certain elements of a given class (rings). This is even more striking with the RHVAE since it tends to stretch the resulting latent space. This effect seems nonetheless mitigated by the use of a multimodal prior such as the VAMP. However, another limitation of \emph{prior-based} methods is that they may sample in locations of the latent space potentially containing very few information (\emph{i.e.} where no data is available). Since the decoder appears to interpolate quite linearly, the classic scheme will generate images which mainly correspond to a superposition of samples (see an example with the red dots in Fig.~\ref{Fig: Shapes Sampling Comparision} and the corresponding samples framed in red). Moreover, there is no way to assess a sample quality before decoding it and assessing visually its relevance. These limitations may lead to a (very) poor representation of the actual data set diversity while presenting quite a few \emph{irrelevant} samples. Impressively, sampling along geodesic paths leads to far more diverse and sharper samples. The new sampling scheme avoids regions that have been poorly prospected so that almost every decoded sample is visually satisfying and accounts for the data set diversity. In Fig.~\ref{Fig: Comparison}, we also compare the models on a \emph{reduced} MNIST \cite{lecun_mnist_1998} data set composed of 120 samples of 3 different classes and a \emph{reduced} FashionMNIST \cite{xiao_fashion-mnist_2017} data set composed again of 120 samples from 3 distinct classes. The models are trained with the same neural network architectures, batch size and learning rate.  An early stopping strategy is adopted and consists in stopping training if the ELBO does not improve for 50 epochs. As discussed earlier, changing the prior may indeed improve the model generation capacity. For instance samples from the VAE with the VAMP prior ($3^{\mathrm{rd}}$ row of Fig.~\ref{Fig: Comparison}) are closer to the training data ($1^{\mathrm{st}}$ row of Fig.~\ref{Fig: Comparison}) than with the Gaussian prior ($2^{\mathrm{nd}}$ and $4^{\mathrm{th}}$ row). The model is for instance able to generate circles when trained with the synthetic data while models using a standard normal prior are not. Nonetheless, a non negligible part of the generated samples are degraded (see saturated images for the \emph{reduced} MNIST data for instance). This aspect is mitigated with the proposed generation method which generates more diverse and sharper samples.

\subsubsection{OASIS Database}

\begin{figure}[ht]
    \centering
    \adjustbox{minipage=3.5em,raise=\dimexpr -2.\height}{ Train (CN)}
    \captionsetup[subfigure]{position=above, labelformat = empty}
    \subfloat{\includegraphics[width=4.2in]{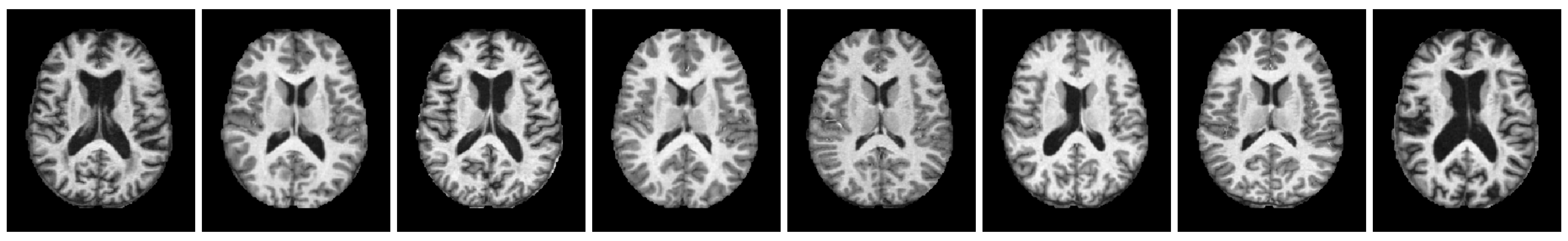}}
    \vfil
    \vspace{-0.17in}
    \adjustbox{minipage=3.5em,raise=\dimexpr -2\height}{ Train (AD)}
    \subfloat{\includegraphics[width=4.2in]{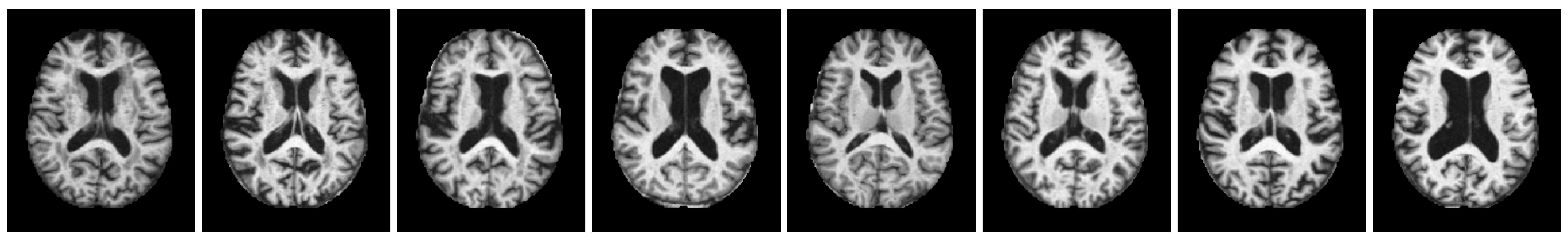}}
    \vfil
    \adjustbox{minipage=3.5em,raise=\dimexpr -2\height}{ VAE (CN)}
    \subfloat{\includegraphics[width=4.2in]{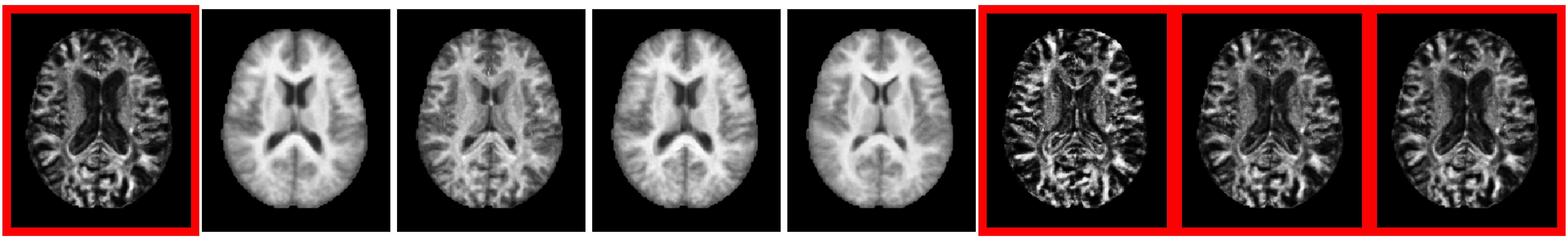}}
    \vfil
    \vspace{-0.17in}
    \adjustbox{minipage=3.5em,raise=\dimexpr -2\height}{ VAE (AD)}
    \subfloat{\includegraphics[width=4.2in]{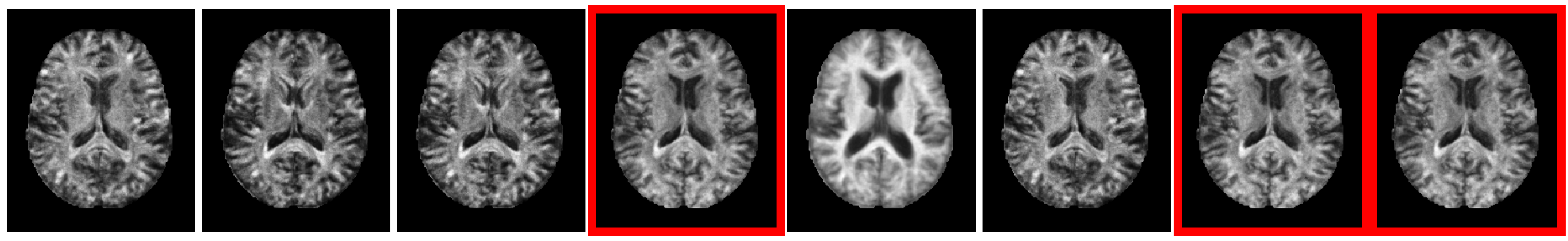}}
    \vfil
    \adjustbox{minipage=3.5em,raise=\dimexpr -2\height}{ Ours (CN)}
    \subfloat{\includegraphics[width=4.2in]{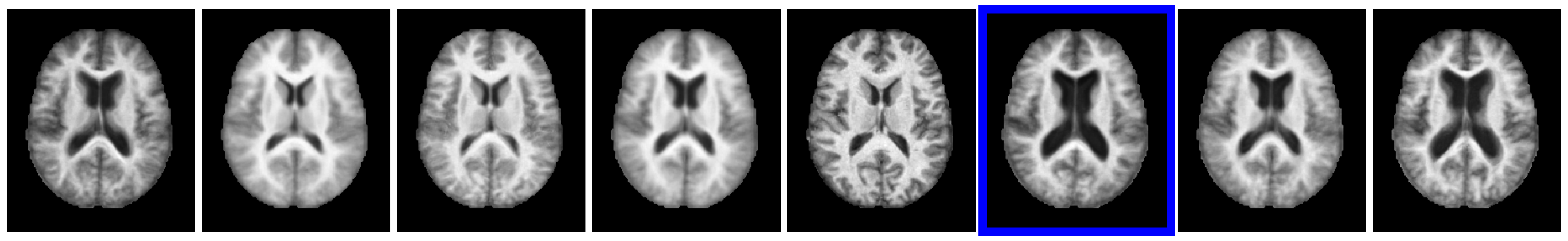}}
    \vfil
    \vspace{-0.17in}
    \adjustbox{minipage=3.5em,raise=\dimexpr -2\height}{ Ours (AD)}
    \subfloat{\includegraphics[width=4.2in]{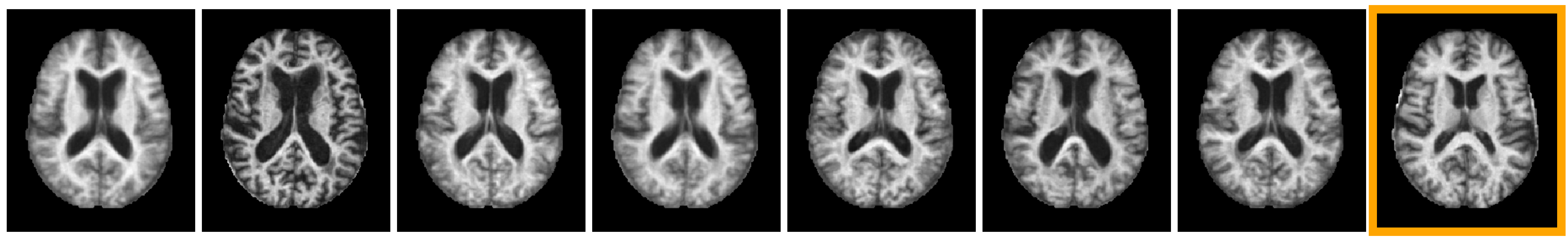}}
    \vfil
    \caption{Generation of CN or AD patients from the OASIS database. Training samples (top), generation with a VAE and normal prior ($2^{\mathrm{nd}}$ row) and with the Riemannian random walk (bottom). Generating using the prior leads to either unrealistic images or similar samples (red frames) while the proposed method generates sharper and more diverse samples. For instance, it is able to generate CN \emph{older} patients (blue frames) or younger AD (orange frames) even though they are under-represented within the training set. }
    \label{Fig: OASIS Generation}
    \end{figure}

The new generation scheme is then assessed on the publicly available OASIS database composed of 416 patients aged 18 to 96, 100 of whom have been diagnosed with very mild to moderate Alzheimer disease (AD). A VAE and a RHVAE are then trained to generate either cognitively normal (CN) or AD patients with the same early stopping criteria as before. Fig.~\ref{Fig: OASIS Generation} shows samples extracted from the training set (top), MRI generated by a vanilla VAE ($2^{\mathrm{nd}}$ row) and images from the Riemannian random walk we propose ($3^{\mathrm{rd}}$ row). For each method, the upper row shows images of patients diagnosed CN while the bottom row presents an AD diagnosis. Again the proposed sampling seems able to generate a wider range of sharp samples while the VAE appears to produce non-realistic degraded images which are very similar (see red frames). For example, the proposed scheme allows us to generate realistic \emph{old}\footnote{An older person is characterised by larger ventricles.} patients with no AD (blue frames) or younger patients with AD (orange frames) even though they are under-represented in the training set. Generating 100 images of OASIS database takes 1 min. with the proposed method and 40 sec.\footnote{Depends on the chains' length (here 200 steps per image).} with Intel Core i7 CPU (6x1.1GHz) and 16 GB RAM.

\section*{Appendix B: Discussion of Remark.~1}
\begin{rem}
    If $\Sigma$ has small enough eigenvalues then Alg.~1 samples from
    \begin{equation}\label{Eq: Likelihoog metric appendix}
    \mathcal{L}(z) = \frac{\rho_S(z)\sqrt{\det \mathbf{G}^{-1}(z)}}{\int \limits_{\mathbb{R}^d}\rho_S(z)\sqrt{\det \mathbf{G}^{-1}(z)}dz}\,,
\end{equation}
where $\rho_S(z)=1$ if $z \in S$, $0$ otherwise, and $S$ is taken as a compact set so that the integral is well defined. 
\end{rem}

If $\Sigma$ has small enough eigenvalues, it means that the initial velocity $v \sim \mathcal{N}(0, \Sigma)$ will have a low magnitude with high probability. In such  a case, we can show with some approximation that the ratio $\alpha$ in the Riemannian random walk is a Hasting-Metropolis ratio with target density given by Eq.~\eqref{Eq: Likelihoog metric appendix}. We recall that the classic Hasting-Metropolis ratio writes
    \[
        \alpha(x, y) = \frac{\pi(y)}{\pi(x)} \cdot \frac{q(x, y)}{q(y, x)}\,,
    \]
    where $\pi$ is the target distribution and $q$ a proposal distribution. In the case of small magnitude velocities, one may show that $q$ is symmetric that is
    \[
        q(x, y) = q(y, x)\,.
    \]
    In our setting, a proposal $\widetilde{z}$ is made by computing the geodesic $\gamma$ starting at $\gamma(0) = z$ with initial velocity $\dot{\gamma}(0) = v$ where $v \sim \mathcal{N}(0, \Sigma)$ and evaluating it at time 1. First, we remark that $\gamma$ is well defined since the Riemnanian manifold $\mathcal{M} = (\mathbb{R^d}, g)$ is \textit{geodesically complete} and $\gamma$ is unique. Moreover, we have that $\overleftarrow{\gamma}$ is the unique geodesic with initial position $\overleftarrow{\gamma}(0) = \widetilde{z} = \gamma(1)$ and initial velocity $\dot{\overleftarrow{\gamma}}(0)=-\dot{\gamma}(1)$ and we have
    \[
        \overleftarrow{\gamma}(t) = \gamma(1-t), \hspace{5mm} \forall~ t \in [0, 1]\,.
    \]
    In the case of small enough initial velocity, a taylor expansion of the exponential may be performed next to $ 0 \in T_z\mathcal{M}$ and consists in approximating geodesic curves with straight lines. That is, for $t \in [0, 1]$
    \[
        \mathrm{Exp}_z(vt) \approx z + vt\,,
    \]
    where $v = \widetilde{z} - z$.
    In such a case we have
    \[
        \dot{\gamma}(t) = v = \dot{\gamma}(0) = \dot{\gamma}(1) = -\dot{\overleftarrow{\gamma}}(0)\,.
    \]
    Moreover, we have on the one hand
    \[
        q(\widetilde{z}, z) = p(\widetilde{z}|z) = p(\mathrm{Exp}_z(v)|z) \simeq \mathcal{N}(z, \Sigma)\,.
          \]
    On the other hand
    \[
        q(z, \widetilde{z}) = p(z|\widetilde{z}) = p(\mathrm{Exp}_{\widetilde{z}}(-v)|\widetilde{z}) \simeq \mathcal{N}(\widetilde{z}, \Sigma) \\  
    \]
    Therefore, \[
        q(\widetilde{z}, z) = q(z, \widetilde{z})\,.
    \]    
    Finally, the ratio $\alpha$ in the Riemannian random walk may be seen as a Hasting-Metropolis ratio where the target density is given by Eq.~\eqref{Eq: Likelihoog metric} and so the algorithm samples from such a distribution.

\clearpage
\section*{Appendix C: Computing the Exponential map}

To compute the exponential map at any given point $p \in \mathcal{M}$ and for any tangent vector $v \in T_p\mathcal{M}$ we rely on the Hamiltonian definition of geodesic curves. First,  for any given $v \in T_p\mathcal{M}$, the linear form:
    \[
        q_v:\Bigg\{\begin{array}{cl}
        T_p\mathcal{M} &\to \mathbb{R}  \\
        u &\to g_p(u, v) 
    \end{array}\,,
    \]
    is called a moment and is a representation of $v$ in the dual space. In short, we may write $q_v(u) = u^{\top}\mathbf{G}v$. Then, the definition of the Hamiltonian follows
    \[
        H(p, q) = \frac{1}{2} g_p^{*}(q, q)\,,
    \]
    where $g_p^{*}$ is the dual metric whose local representation is given by $\mathbf{G}^{-1}(p)$, the inverse of the metric tensor. Finally, all along geodesic curves the following equations hold
    \begin{equation}\label{Eq: Hamiltonian Eq}
     \frac{\partial p}{\partial t} = \frac{\partial H}{\partial q}, \hspace{5mm}
     \frac{\partial q}{\partial t}  = - \frac{\partial H}{\partial p}\,.
    \end{equation}
Such a system of differential equations may be integrated pretty straight forwardly using simple numerical schemes such as the second order Runge Kutta integration method and Alg.~\ref{Alg: Exponential map} as in \cite{louis_computational_2019}. Noteworthy is the fact that such an algorithm only involves one metric tensor inversion at initialization to recover the initial moment from the initial velocity. Moreover, it involves closed form operations since the inverse metric tensor $\mathbf{G}^{-1}$ is known (in our case) and so the gradients in Eq.~\eqref{Eq: Hamiltonian Eq} can be easily computed.

\begin{algorithm}[ht]
    \caption{Computing the Exponential map}
    \label{Alg: Exponential map}
   \begin{algorithmic}
      \STATE{\bfseries Input:} $z_0 \in \mathcal{M}$, $v \in T_{z_0}\mathcal{M}$ and $T$
      \STATE $q \leftarrow \mathbf{G}\cdot v$
      \STATE $dt \leftarrow \frac{1}{T}$
       \FOR{$t=1 \rightarrow T$}
       \STATE $p_{t+ \frac{1}{2}} \leftarrow p_t + \frac{1}{2} \cdot dt \cdot\nabla_q H(p_t, q_t)$
       \STATE $q_{t+ \frac{1}{2}} \leftarrow q_t - \frac{1}{2} \cdot dt \cdot \nabla_p H(p_t, q_t)$
       \STATE $p_{t+1} \leftarrow p_t + dt \cdot \nabla_q H(p_{t+ \frac{1}{2}}, q_{t+ \frac{1}{2}})$
       \STATE $q_{t+ 1} \leftarrow q_t - dt \cdot \nabla_p H(p_{t+ \frac{1}{2}}, q_{t+ \frac{1}{2}})$
   \ENDFOR
   \STATE {\bfseries Return $p_T$}
   \end{algorithmic}
\end{algorithm}

\clearpage
\section*{Appendix D: VAEs Parameters Setting}

Table.~\ref{Table: Hyper-parameters} summarizes the main hyper-parameters we use to perform the experiments presented in the paper while Table.~\ref{Tab: Neural Networks Architectures} shows the neural networks architectures employed. As to training parameters, we use a Adam optimizer \cite{kingma_adam_2014} with a learning of $10^{-3}$. For the augmentation experiments, we stop training if the ELBO does not improve for 20 epochs for all data sets except for OASIS where the learning rate is decreased to $10^{-4}$ and training is stopped if the ELBO does not improve for 50 epochs. 
\begin{table}[!ht]
    \centering
    \scriptsize
    \caption{RHVAE parameters for each data set.}
    \label{Table: Hyper-parameters}
    \begin{sc}
    \begin{tabular}{c|cccccc}
    \toprule
         Data sets & \multicolumn{6}{c}{Parameters} \\
         & $d^{*}$  & $n_{\mathrm{lf}}$ & $\varepsilon_{\mathrm{lf}}$ & $T$ & $\lambda$ & $\sqrt{\beta_0}$   \\
        \midrule
        Synthetic                   & 2 & 3 & $10^{-2}$ & 0.8 & $10^{-3}$ & 0.3 \\
        \textit{reduced} Fashion   & 2 & 3 & $10^{-2}$ & 0.8 & $10^{-3}$ & 0.3 \\
        MNIST (bal.)      & 2 & 3 & $10^{-2}$ & 0.8 & $10^{-3}$ & 0.3 \\
        MNIST (unbal.)      & 2 & 3 & $10^{-2}$ & 0.8 & $10^{-3}$ & 0.3 \\
        EMNIST      & 2 & 3 & $10^{-2}$ & 0.8 & $10^{-3}$ & 0.3 \\
         OASIS                      & 2 & 3 & $10^{-3}$ & 0.8 & $10^{-2}$ & 0.3 \\
        \bottomrule
        \multicolumn{7}{l}{\tiny{* Latent space dimension (same for VAE and VAMP-VAE)}} 
    \end{tabular}
\end{sc}
\end{table}

\begin{table}[!ht]
\caption{Neural networks architectures of the VAE, VAMP-VAE and RHVAE for each data set. The \emph{encoder} and \emph{decoder} are the same for all models.}
\label{Tab: Neural Networks Architectures}
    \centering
    \scriptsize
    \begin{sc}
    \begin{tabular}{cccc}
    \toprule
    \multicolumn{4}{c}{Synthetic, MNIST \& Fashion}\\
    \midrule
    \midrule
    Net & Layer 1 & Layer 2 & Layer 3\\
    \midrule
        $\mu_{\phi}^* $      &  \multirow{2}{*}{($D$, 400, relu)}& (400, $d$, lin.)& -   \\
        $\Sigma_{\phi}^*$    &                                 & (400, $d$, lin.) & -  \\
         \midrule
        $ \pi_{\theta}^* $   &  ($d$, 400, relu)                 & (400, $D$, sig.) &-    \\
         \midrule
        $L_{\psi}$ \tiny{(diag.)}  &  \multirow{2}{*}{($D$, 400, relu)}& (400, $d$, lin.) & -    \\
        $L_{\psi}$ \tiny{(low.)}    &                                 & (400, $ \frac{d(d-1)}{2}$, lin.)& -\\
        \midrule
        \midrule
        \multicolumn{4}{c}{OASIS}\\
        \midrule
        \midrule
        $\mu_{\phi}^* $      &  \multirow{2}{*}{($D$, 1k, relu)} & \multirow{2}{*}{(1k, 400, relu)} & (400, $d$, lin.)   \\
        $\Sigma_{\phi}^*$    &                                    & & (400, $d$, lin.) \\
         \midrule
        $ \pi_{\theta}^* $   &  ($d$, 400, relu) & (400, 1k, relu)                & (1k, $D$, sig.)   \\
         \midrule
        $L_{\psi}$ \tiny{(diag.)}  &  \multirow{2}{*}{($D$, 400, relu)}& (400, $d$, lin.)&  -  \\
        $L_{\psi}$ \tiny{(low.)}    &                                 & (400, \tiny{$ \frac{d(d-1)}{2}$}, lin.) & - \\
        \bottomrule
    \multicolumn{3}{l}{\tiny{* Same for all VAE models}}
    \end{tabular}
\end{sc}
\end{table}

\clearpage

\section*{Appendix E: Classifier Parameter Setting}

As to the models used as benchmark for data augmentation, the DenseNet implementation we use is the one in \cite{amos_bamosdensenetpytorch_2020} with a \emph{growth rate} equals to 10, \emph{depth} of 20 and 0.5 \emph{reduction} and is trained with a learning rate of $10^{-3}$. For OASIS, the MLP has 400 hidden units and relu activation function and the CNN is as follows

\begin{table}[!ht]
\caption{CNN classifier architecture used. Each convolutional block has a padding of 1.}
\label{Tab: CNN Architectures}
    \centering
    \scriptsize
    \begin{sc}
    \begin{tabular}{cc}
    \toprule
    Layer & Architectures \\
    \midrule
    input & (1, 208, 176) \\
    \midrule
    \multirow{4}{*}{Layer 1} & Conv2D(1, 8, kernel=(3, 3), stride=1) \\
                             & Batch normalization \\
                             & LeakyRelu \\
                             & Maxpool (2, 2, stride=2)\\
    \midrule
    \multirow{4}{*}{Layer 2} & Conv2D(8, 16, kernel=(3, 3), stride=1) \\
                             & Batch normalization \\
                             & LeakyRelu \\
                             & Maxpool (2, 2, stride=2)\\
    \midrule
    \multirow{4}{*}{Layer 3} & Conv2D(16, 32, kernel=(3, 3), stride=2) \\
                             & Batch normalization \\
                             & LeakyRelu \\
                             & Maxpool (2, 2, stride=2)\\
    \midrule
    \multirow{4}{*}{Layer 4} & Conv2D(32, 64, kernel=(3, 3), stride=2) \\
                             & Batch normalization \\
                             & LeakyRelu \\
                             & Maxpool (2, 2, stride=2)\\
    \midrule
    \multirow{2}{*}{Layer 5} & MLP(256, 100) \\
                             & Relu \\
    \midrule
    \multirow{2}{*}{Layer 6} & MLP(100, 2) \\
                             & Log Softmax \\
    \bottomrule
    \end{tabular}
\end{sc}
\end{table}

For the toy data, the DenseNet is trained until the loss does not improve on the validation set for 50 epochs. On OASIS, we make a random search on the learning rate for each model chosen in the range $ [10^{-1}, 10^{-2}, 10^{-3}, 10^{-4}, 10^{-5}, 10^{-6}]$. The model is trained on 5 independent runs with the same learning rate and keep the model achieving the best mean balanced accuracy on the validation set. For the CNN and MLP, we stop training if the validation loss does not improve for 20 epochs and for the densenet training is stopped if no improvement is observed on the validation loss for 10 epochs. Each model is trained with an Adam optimizer.

%
%
%
%

\end{document}